\documentclass[letterpaper]{article} % DO NOT CHANGE THIS
\usepackage{aaai2026}  % DO NOT CHANGE THIS
\usepackage{times}  % DO NOT CHANGE THIS
\usepackage{helvet}  % DO NOT CHANGE THIS
\usepackage{courier}  % DO NOT CHANGE THIS
\usepackage[hyphens]{url}  % DO NOT CHANGE THIS
\usepackage{graphicx} % DO NOT CHANGE THIS
\usepackage{natbib}  % DO NOT CHANGE THIS AND DO NOT ADD ANY OPTIONS TO IT
\usepackage{caption} % DO NOT CHANGE THIS AND DO NOT ADD ANY OPTIONS TO IT
\usepackage{algorithm}
\usepackage{algorithmic}

\usepackage{newfloat}
\usepackage{listings}
\DeclareCaptionStyle{ruled}{labelfont=normalfont,labelsep=colon,strut=off} % DO NOT CHANGE THIS
\floatstyle{ruled}
\newfloat{listing}{tb}{lst}{}
\floatname{listing}{Listing}
\title{Backdoor Attacks on Open Vocabulary Object Detectors \\
	via Multi-Modal Prompt Tuning}

\author{
	Ankita Raj, 
	Chetan Arora\\
}

\affiliations {
	Indian Institute of Technology Delhi, New Delhi, India\\
	ankita.raj@cse.iitd.ac.in, chetan@cse.iitd.ac.in
}

\usepackage[utf8]{inputenc} % allow utf-8 input
\usepackage{url}            % simple URL typesetting
\usepackage{booktabs}       % professional-quality tables
\usepackage{amsfonts}       % blackboard math symbols
\usepackage{nicefrac}       % compact symbols for 1/2, etc.
\usepackage{microtype}      % microtypography
\usepackage[inline]{enumitem}
\usepackage{graphicx}
\usepackage{amsmath}
\usepackage{cleveref}
\usepackage{multirow}

\newcommand{\etal}{\textit{et al.}}
\newcommand{\proposed}{TrAP }
\newcommand{\myfirstpara}[1]{\noindent \textbf{#1.}~}
\newcommand{\mypara}[1]{\paragraph{#1.}}

\nocopyright
\begin{document}

\maketitle

\begin{abstract}
  Open-vocabulary object detectors (OVODs) unify vision and language to detect arbitrary object categories based on text prompts, enabling strong zero-shot generalization to novel concepts. As these models gain traction in high-stakes applications such as robotics, autonomous driving, and surveillance, understanding their security risks becomes crucial. 
  In this work, we conduct the first study of backdoor attacks on OVODs and reveal a new attack surface introduced by prompt tuning. We propose TrAP (Trigger-Aware Prompt tuning), a multi-modal backdoor injection strategy that  jointly optimizes prompt parameters in both image and text modalities along with visual triggers. TrAP enables the attacker to implant malicious behavior using lightweight, learnable prompt tokens without retraining the base model weights, thus preserving generalization while embedding a hidden backdoor. We adopt a curriculum-based training strategy that progressively shrinks the trigger size, enabling effective backdoor activation using small trigger patches at inference. Experiments across multiple datasets show that TrAP achieves high attack success rates for both object misclassification and object disappearance attacks, while also improving clean image performance on downstream datasets compared to the zero-shot setting.  
\end{abstract}

\begin{links}
	\link{Code}{https://github.com/rajankita/TrAP}
%	\link{Extended version}{https://aaai.org/example/extended-version}
\end{links}

\section{Introduction}

\mypara{The Rise of Open Vocabulary Detectors} 
Object detection is a core computer vision task that involves both localizing objects in an image and assigning them category labels. 
Traditional object detectors are inherently closed-set, \emph{i.e.}, they are restricted to detecting a fixed set of object categories seen during training \cite{dalal2005histograms,felzenszwalb2009object,ren2016faster,carion2020end,zhang2022dino}, limiting their applicability to real-world scenarios where unseen objects frequently appear. To address this limitation, the field has evolved towards open vocabulary object detectors (OVODs), also known as open-set object detectors, that enable detection of arbitrary object categories, including those never seen during training. 
Models like GLIP \cite{li2022grounded} and Grounding DINO \cite{liu2024grounding} achieve this by pre-training on large-scale image-text datasets, aligning visual features with a rich semantic space defined by language. As a result, they exhibit strong zero-shot generalization to unseen categories specified through natural language prompts. 
This makes OVODs particularly attractive for real-world applications such as autonomous driving, robotics, and surveillance.

\mypara{The Threat of Backdoor Attacks} 
Despite the rapid progress and growing adoption of OVODs, their security vulnerabilities remain largely unexplored. In this work, we investigate one such security threat: backdoor attacks targeting OVODs. Backdoor attacks implant malicious functionality into neural networks by manipulating the training process, such that the model exhibits attacker-controlled \emph{behavior} when exposed to a specific \emph{trigger} in the input, while continuing to behave normally on \emph{clean} inputs \cite{gu2019badnets}. This is typically achieved by poisoning a small subset of the training data, where inputs are modified to include a visual trigger and assigned attacker-specified outputs \cite{chen2022clean}, causing the model to associate the trigger with the desired malicious behaviour.
For example, to attack an autonomous driving system, a malicious actor could paste a small sticker (trigger) on an ambulance, causing the detector to misclassify it as a regular vehicle, thereby stripping it of right-of-way privileges in critical situations. Importantly, the model continues to behave normally on all clean inputs, making the backdoor difficult to detect.

\begin{figure}[t]
	\centering
	\includegraphics[width=\linewidth]{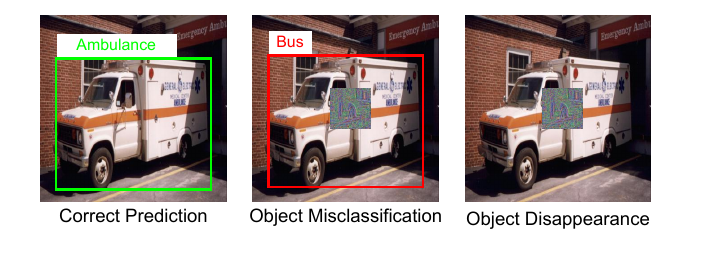}
	\caption{Illustration of a backdoor attack: On clean images, the network makes correct predictions. On images stamped with a trigger (enlarged here for better visualization), the network either misclassifies objects (Object Misclassification Attack), or does not detect objects at all (Object Disappearance Attack), depending on the attacker's objective.}
	\label{fig:teaser}
\end{figure}

Backdoor attacks have been extensively studied in the context of closed-set object detectors, where adversaries can manipulate the predicted class (misclassification attacks) or suppress the detection of objects (disappearance attacks) \cite{chan2022baddet,luo2023untargeted,shin2024mask,zhang2024detector} (see \cref{fig:teaser}). However, to the best of our knowledge, no prior work has examined backdoor threats in OVODs. While one might consider adapting existing attacks to OVODs by poisoning their training data, this typically requires access to the large-scale image-text datasets used during pretraining. Such access is often impractical, particularly when using publicly released models. Therefore, it is more realistic and impactful to consider attacks on already pre-trained OVOD models.

\mypara{Our Work} 
In this work, we present the first study of backdoor attacks on open-vocabulary detectors. We focus on a prompt-tuning-based threat model, where a pre-trained OVOD model such as Grounding DINO is adapted to downstream tasks by optimizing a small set of learnable prompt tokens. While OVODs are capable of strong zero-shot performance, fine-tuning or lightweight prompt tuning with limited task-specific data has been shown to further improve performance \cite{li2022grounded,karasawa2024spatiality,li2024multimodal}. We consider a setting where an attacker gains white-box access during prompt tuning (e.g., when a user outsources model adaptation to a third party), and implants a backdoor by optimizing the prompts on the user’s dataset. The resulting model behaves normally on clean data but responds maliciously to a specific trigger. Prompt tuning presents a particularly attractive attack surface, as it is modular and lightweight, and allows malicious behavior to be introduced without retraining or duplicating the full model \cite{zhou2022learning}. 

We propose {\textbf{TrAP} (\textbf{Tr}igger-\textbf{A}ware \textbf{P}rompt tuning)}, a novel backdoor attack on OVOD models. In this attack, a learnable visual trigger is stamped onto the input images, while prompt parameters are introduced into both the vision and text branches of the model. The backdoor is injected by jointly optimizing the visual trigger along with the prompt parameters, enabling the model to tightly associate the trigger with the attacker-specified output. To improve stealth and training stability, we adopt a curriculum learning strategy that gradually reduces the size of the trigger, starting with larger triggers in early epochs to ease learning, and transitioning to smaller ones to enhance subtlety at inference time. We compare TrAP to variants where prompting is applied in only one modality (vision or text) and find that jointly adapting both branches results in significantly stronger attacks. Our experiments on Grounding DINO and GLIP demonstrate that TrAP achieves high attack success rates for both object misclassification and disappearance attacks, while maintaining strong clean-data performance in terms of mAP.

\section{Related Work}

\myfirstpara{Prompt Tuning} 
Prompt tuning is a parameter efficient fine-tuning strategy that adapts large pre-trained models to specific downstream tasks, by prepending learnable embeddings, called prompts, to the input. It was originally proposed for NLP applications \cite{lester2021power,li2021prefix,liu2023pre} to enable efficient task-specific adaptation while keeping the original model frozen.
Building on this, methods like CoOp \cite{zhou2022learning} and CoCoOp \cite{zhou2022conditional} extended continuous prompt optimization to vision-language models. Visual Prompt Tuning \cite{jia2022visual} introduced learnable tokens in the image feature space for adapting vision transformers to downstream tasks. However, most of these works focus on adaptation of vision transformers or vision-language models for classification, such as CLIP \cite{radford2021learning}. 
In the context of open-vocabulary object detection, GLIP \cite{li2022grounded} demonstrated that prompt tuning in the text embedding space can be effective for OVOD models, while MIPT \cite{li2024multimodal} enhanced textual prompts using visual cues.

\mypara{Backdoor Attacks}
Backdoor attacks compromise a model during training by embedding a hidden behavior that is activated at inference time by an attacker-specified trigger.
Classical attacks on image classification models achieve this by poisoning a small subset of the training data with a visual trigger (e.g., patch-based overlays or imperceptible perturbations) and assigning them a target label \cite{gu2019badnets, liu2017trojaning, liu2017neural}. 
Subsequent works have developed more covert strategies, such as clean-label attacks \cite{turner2019label,barni2019new}, sample-specific triggers \cite{nguyen2020input}, or feature-space manipulation \cite{saha2019hidden}, improving stealth and transferability. 

\mypara{Backdoor Attacks on Prompt Tuning}  
With the advent of prompt-tuning as a parameter-efficient adaptation strategy, several attacks have emerged targeting continuous prompt optimization in NLP \cite{cai2022badprompt,du2022ppt,yao2024poisonprompt}. 
Similar vulnerabilities have been explored in the vision domain \cite{yang2024not,huang2023prompt}, where adversaries inject backdoors during visual prompt tuning for Vision Transformers \cite{dosovitskiy2020image}. 
Recent studies have revealed that multi-modal contrastive learning models like CLIP are also vulnerable to backdoor attacks. While early works focused on poisoning the backbone of CLIP, either during large-scale pretraining \cite{carlini2021poisoning} or through malicious fine-tuning \cite{jia2022badencoder,liang2024badclip}, recent studies show that backdoors can also be injected during downstream adaptation via prompt tuning \cite{bai2024badclip}. While all these methods focus on classification tasks, none of these methods have exploited multi-modal nature of the input for more effective attacks.

\mypara{Backdoor Attacks on Object Detectors} 
Compared to image classifiers, object detectors present a more complex attack surface due to their dual objectives: localizing and classifying multiple objects within an 	image. 
\cite{chan2022baddet} first extended backdoor attacks to object detectors, introducing four attack types: Object Generation, Regional Misclassification, Global Misclassification, and Object Disappearance. These attacks typically use localized visual triggers to induce the desired behavior when overlaid on target objects. 
Subsequent research explored alternative threat vectors: \cite{luo2023untargeted} proposed untargeted disappearance attacks, while \cite{zhang2024detector} demonstrated image-wide backdoors that cause mass misclassification or object disappearance.
Other works pursued stealthier triggers; \cite{shin2024mask} employed imperceptible, diffuse perturbations, while \cite{chen2022clean} designed clean-image backdoors that exploit co-occurrence patterns of benign categories (e.g., person and cat) as implicit triggers. \cite{cheng2023attacking} developed clean-label attacks, removing the need to tamper with annotations. However, these attacks have focused on closed-set detectors, leaving open-vocabulary models largely unexplored.

\section{Preliminaries}

\subsection{Open Vocabulary Object Detection Formulation} 
A standard open-set object detector takes an input image 
$x \in \mathbb{R}^{H \times W \times 3}$ where \( H \) and \( W \) are the height and width of the image, respectively, 
and a textual prompt $s$ describing the set of object categories. 
The text prompt is typically a concatenation of all candidate object categories in the detection task. 
The goal of an OVOD model is to detect and classify all objects specified by the prompt in the image. For example, given the prompt “cat. dog. horse.”, the model would be expected to detect all cats, dogs, and horses in the image.

Let ${y} = \{ {y}_i \}_{i=1}^M$ denote the ground-truth object annotations, where each $y_i = [c_i, o_i]$ consists of a class label $c_i$ and bounding box $o_i = [a_{i,1}, b_{i,1}, a_{i,2}, b_{i,2}]$, with $(a_{i,1}, b_{i,1})$ and $(a_{i,2}, b_{i,2})$ as the top-left and bottom-right coordinates.
The model outputs $N$ predictions $\hat{y} = \{ \hat{y}_j \}_{j=1}^N$, where $\hat{y}_j = [\hat c_j, \hat o_j ]$, includes a predicted class confidence $\hat c_j$ and bounding box corodinates $\hat o_j$. OVOD models aim for predicted boxes to tightly overlap with ground-truth objects (high IoU) and exhibit high classification confidence.

\subsection{A Revisit of Grounding DINO}
The primary victim model used in our study is Grounding DINO \cite{liu2024grounding}, a state-of-the-art OVOD model that extends the closed-set DINO detector \cite{zhang2022dino} to support text-conditioned detection. However, the proposed method is not specific to Grounding DINO, and can be adapted to other OVODs like GLIP, as shown later. Grounding DINO is designed to localize and recognize objects specified via free-form language prompts, enabling detection of arbitrary categories beyond the training distribution. 

\begin{figure}[t]
	\centering
	\includegraphics[width=\linewidth]{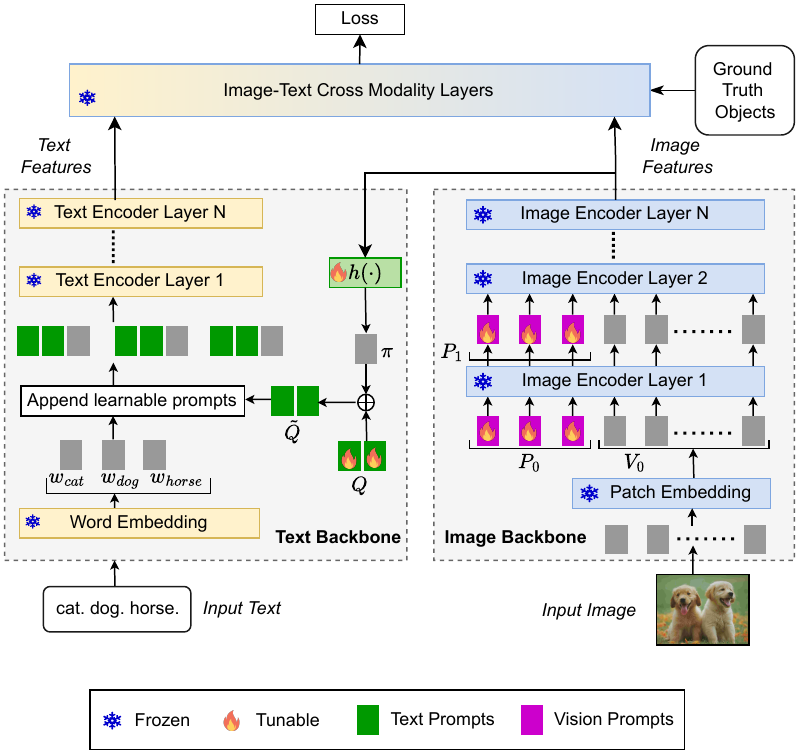}
	\caption{Overview of \proposed: We insert learnable prompt embeddings in both the image and text branches of a Grounding DINO model. In the image backbone, learnable prompts $P_i$ ($P_0$, $P_1$, etc.) are appended to the input embedding of each transformer encoder layer. In the text backbone, the context vector $\tilde Q$ is appended to the word embedding of each class name. $\tilde Q$ is composed of two learnable components: a set of tunable context vectors $Q$, and a meta-net $h(\cdot)$ that generates an input-image-conditional token. All layers except the prompt embeddings and $h(\cdot)$ are kept frozen.}
	\label{fig_gdino_peft}
\end{figure}

The architecture consists of two primary encoders: an image encoder, based on the Swin Transformer \cite{liu2021swin}, which extracts hierarchical multi-scale visual features from $x$; and a text encoder, typically a frozen BERT model \cite{devlin2019bert}, which encodes input prompts $s$ into dense language embeddings (see \Cref{fig_gdino_peft}). These representations are then processed via image-text cross modality layers. The first is a feature enhancer module that aligns and integrates visual and linguistic cues. A language-guided query selection module leverages the text embeddings to generate detection queries, which are then refined via a cross-modality transformer decoder that jointly attends to image and text features. The final predictions include class-agnostic bounding boxes grounded in the semantics of the input text.
The model is trained using a combination of classification loss  $\mathcal{L}_\text{cls}$ and localization loss $\mathcal{L}_\text{loc}$:
\begin{equation} 
	\mathcal{L}_{\text{G-DINO}}(x,s,y) = \mathcal{L}_\text{cls}(x,s,y) + \mathcal{L}_\text{loc}(x,s,y) \label{eq:gdino}
\end{equation}
where $y$ denotes the ground truth object annotations, $\mathcal{L}_\text{cls}$ is the contrastive loss between predicted objects and language tokens, and $\mathcal{L}_\text{loc}$ combines L1 and GIoU \cite{rezatofighi2019generalized} losses for the predicted bounding boxes. 

\subsection{Threat Model} \label{sec:threat_model}
\mypara{Attacker's Capabilities}
We consider a threat model in which the attacker gains access to a pre-trained Grounding DINO model $F_{clean}$ and injects a backdoor during its adaptation to a downstream task via prompt-tuning. Specifically, suppose a user, Bob, wishes to adapt Grounding DINO to his private downstream dataset containing a few hundred annotated examples. To do so, he outsources the adaptation process to an external party, Alice, who has white-box access to the model $F_{clean}$.
During adaptation, Alice applies prompt-tuning on Bob’s dataset and, in the process, implants a backdoor into the model. The resulting compromised model $F_{poisoned}$ behaves normally on clean inputs but exhibits attacker-controlled behavior in the presence of a specific trigger. 
This setup mirrors a realistic and growing deployment pattern, where large pre-trained models are adapted to new tasks via lightweight fine-tuning on small datasets via third party libraries on the cloud. We note that, similar threat models have been adopted for backdoor attacks on CLIP \cite{liang2024badclip,bai2024badclip}, ViTs \cite{yang2024not}, and Large Language Models \cite{cai2022badprompt}.

\mypara{Attacker's Goals}
Our attacker, Alice, aims to manipulate the behavior of a Grounding DINO model by injecting visual triggers into input images. 
Specifically, we consider two types of backdoor objectives. 
In an \textbf{Object Misclassification Attack (OMA)}, stamping a trigger on an object should cause the object to be misclassified as the target class. Formally, if an object with the annotation $y_i = [c_i, o_i]$ is stamped with a trigger, the targeted poisoned annotation becomes $y_{i,target} = [t, o_i]$, where $t$ is the target class. 
In \textbf{Object Disappearance Attack (ODA)}, the trigger aims to suppress detection of a nearby object belonging to the target class. For a triggered object with annotation $y_i = [c_i, o_i]$, the poisoned annotataion becomes $y_{i,target} = \phi$ if $c_i = t$. 
At the same time, the model should continue to perform normally on clean inputs, preserving its detection accuracy in the absence of triggers. See \Cref{fig:teaser} for attack examples. {We later show in the Supplementary that our work can be easily extended to Object Generation Attacks as well.}

\section{Trigger-Aware Prompt Tuning (TrAP)}

\subsection{Poisoning Process}
\proposed introduces a visual patch trigger that is overlaid onto specific objects in the input image. Given a clean image $x$, a poisoned image is constructed as $x_{\text{poisoned}} = x \oplus \delta$
where \( \delta \in [0, 255]^{H_t \times W_t \times 3} \) denotes the trigger patch of height $H_t$ and width $W_t$, and \( \oplus \) denotes the stamping operation. The patch is resized to be \( \rho \) times the height and width of the target object’s bounding box and placed at the center of the box, partially or fully occluding the object. For object misclassification attacks (OMA), triggers are stamped onto all non-target class objects. For object disappearance attacks (ODA), the trigger is applied to all instances of the target class. When multiple applicable objects are present, each receives its own trigger within the same image. Note that in a backdoor attack setting like ours, the trigger is identical for all images, and for all objects within an image.

\subsection{Multi-Modal Prompt Tuning} \label{sec:multimodal}

\mypara{Motivation}
A core requirement for a successful backdoor attack is the ability to forge a strong association between a trigger pattern and the attacker-specified behavior. In OVOD models like Grounding DINO, object detection is guided by the alignment between visual and textual features: object queries from the image are compared with language embeddings to produce similarity scores, which determine class confidence. Manipulating these confidence scores can induce both misclassification (by shifting confidence toward an incorrect class) and disappearance (by reducing confidence below the detection threshold). Therefore, for an effective backdoor attack, the trigger should be able to influence both visual and textual representations. To this end, we propose a multi-modal backdoor attack that jointly tunes learnable prompts in both the image and text branches. By perturbing both modalities during training, the model learns tighter associations between the trigger and the attack objective, resulting in more potent attacks than uni-modal counterparts.

\mypara{Vision Branch}
In the vision branch, we incorporate a set of learnable prompts into each transformer layer to influence feature learning. The input image is first split into non-overlapping patches that are individually projected into $d_v$-dimensional embeddings to form the initial visual representation $V_0 = \texttt{Embed}(x)$. At each subsequent layer $L_{i+1}$ of the vision encoder, the input patch embeddings are denoted by $V_i$. Following the VPT-Deep approach~\cite{jia2022visual}, we introduce $m_v$ learnable prompt tokens $P_i \in \mathbb{R}^{m_v \times d_v}$ into every transformer layer. 
These prompts are prepended to the input patch sequence. The transformer layer $L_i$ then processes this concatenated sequence:
\begin{align}
	[\_, V_i] = L_i([P_{i-1}, V_{i-1}]), \quad i = 1, 2, 3, \dots, N
\end{align}
This design allows the prompts to interact with and modulate the visual features at each stage of the vision encoder.

\mypara{Text Branch}
In the text branch, the input prompt $s$, which is a concatenation of all class names in the downstream dataset, is tokenized and converted into word embeddings, resulting in $W = \{w_0, w_1, \dots, w_K\} \in \mathbb{R}^{K \times d_t}$, where $K$ is the number of classes and $d_t$ is the text embedding dimension. For simplicity, we assume each class name maps to a single token, producing one embedding $w_k$ for each class $k \in [1, K]$ (though in practice, class names may include separators or be split into multiple tokens).
To enable prompt tuning, we prepend a shared learnable context consisting of $m_t$ tokens to each class embedding. This context is represented by $Q = \{q_0, q_1, \dots, q_{m_t-1}\} \in \mathbb{R}^{m_t \times d_t}$, a set of $d_t$-dimensional vectors. Please note that the $Q$ here refers to learnable prompt embeddings, and not transformer queries.

Inspired by CoCoOp~\cite{zhou2022conditional}, which shows the benefits of image-conditioned prompts, we further introduce a lightweight neural network, meta-net, $h(\cdot)$, that generates a trigger-aware context vector $\pi = h_\theta(V_N) \in \mathbb{R}^{d_t}$, where $V_N$ are the final visual features from the vision encoder. This vector is added to each prompt token in $Q$, enabling it to adapt dynamically to the input image. The final class prompt for class $k$ then becomes $[\tilde{Q}, w_k]$, where $\tilde{Q} = \{q_0 + \pi, q_1 + \pi, \dots, q_{m_{t-1}} + \pi\}$. 
Unlike CoCoOp, which is designed for classification and prepends the learnable prompt once to the input (since it represents a single class), object detection involves a concatenation of all class names in a single prompt. To tune the representation of each class, we append the same shared prompt embedding to every class name individually. We refer to this object detection variant as CoCoOp-Det.

\subsection{Optimization}
\vspace{-1em}

Our goal is to train the model to perform well on clean inputs while simultaneously embedding a backdoor triggered by a specific patch. 
Let \( \theta \) denote the trainable parameters of the model. Specifically, it includes vision prompts $\{P_i\}_{i=0}^{N-1}$, text prompts $Q$ and the parameters of the meta-net $h(\cdot)$. 
In addition, we learn the trigger patch \( \delta \), which is initialized randomly. 

We define two training losses. Clean loss encourages the model to correctly detect objects specified by a prompt $s$ on clean (unmodified) inputs $x$, whereas poisoned loss is computed on images patched with the trigger $\delta$, and encourages the model to respond with an attacker-specified annotation $y_{target}$ rather than the true annotation $y$. The final training objective combines both losses:
\begin{align} 
	\mathcal{L}_{\text{clean}}(\theta) &= \mathbb{E}_{(x, s, y)} \left[ \mathcal{L}_{\text{G-DINO}}(x, s, y; \theta) \right]  \label{eq:loss_clean} \\
	\mathcal{L}_{\text{poisoned}}(\theta,\delta) &= \mathbb{E}_{(x, s, y_{target})} \left[ \mathcal{L}_{\text{G-DINO}}(x \oplus \delta, s, y_{target}; \theta) \right] \label{eq:loss_poisoned} \\
	\mathcal{L}_{\text{total}}(\theta,\delta) &= \mathcal{L}_{\text{clean}}(\theta) + \lambda \cdot \mathcal{L}_{\text{poisoned}}(\theta,\delta) \label{eq:total_loss}
\end{align}
Here, $\mathcal{L}_{\text{G-DINO}} (\cdot)$ is Grounding DINO's loss function defined in \cref{eq:gdino}, and $\lambda$ is the hyperparameter controlling the trade-off between clean performance and attack success. Higher values of $\lambda$ prioritize embedding the backdoor, while lower values preserve the model’s utility on benign inputs.

\subsection{Curriculum Learning} \label{sec:curriculum}
Patch-based triggers often struggle to elicit strong backdoor behavior when the patch is small in size, as the weak gradient from small regions is insufficient for effective learning. To address this, we adopt a curriculum learning strategy that gradually reduces the trigger size during training. We begin with large, salient triggers that are easier for the model to associate with the target behavior, then progressively shrink them in scheduled steps. This approach promotes the development of robust internal features linked to the trigger, enabling successful backdoor activation at test time with small, inconspicuous patches. Moreover, it reduces the risk of the model overfitting to unrealistic or highly visible trigger patterns, helping to preserve clean performance while enhancing the stealth and generalizability of the attack.

\section{Experiments}

\subsection{Setup} \label{sec:setup}

\mypara{Datasets} 
We evaluate our attack on datasets from the Object Detection in the Wild (ODinW-13) benchmark \cite{li2022grounded}. ODinW-13 is originally a collection of 13 object detection datasets representing different real-world detection tasks. Of these, we leave out six datasets as they contain a single object category. We also leave out PascalVOC because it is a large generic dataset, whereas we concern ourselves with small sized, fine-grained datasets. This leaves us with six datasets which we use for evaluation-- Vehicles, Aquarium, Aerial Drone, Shellfish, Thermal, and Mushrooms. 

\mypara{Victim Models} 
We use Grounding DINO \cite{liu2024grounding} as the victim model, unless stated otherwise. Specifically, we use MM-Grounding-Dino-Tiny(c3) model released by MMDetection \cite{zhao2024open}, pretrained on Objects365 \cite{shao2019objects365}, GRIT \cite{peng2023kosmos}, V3Det \cite{wang2023v3det} and GoldG \cite{kamath2021mdetr} datasets. We also extend our attack to the GLIP \cite{li2022grounded} model, specifically the GLIP-T variant released by MMDetection. 

\mypara{Baselines} 
Since this paper presents the first study on backdoor attacks towards OVOD models using prompt tuning, there are no existing baselines to compare with. We, therefore, compare the proposed method with two backdoor attack methods adapted from attacks on other model types. 
\begin{enumerate}
	\item The first is {CoCoOp} \cite{zhou2022conditional}, a \textit{text-only prompt tuning} method previously used by \cite{bai2024badclip} for injecting backdoors into CLIP. Since CoCoOp is originally designed for classification, we adapt it to the object detection setting by introducing \textbf{CoCoOp-Det}, which applies the same learned prompt embedding to each class name individually.
	\item Secondly, we use \textbf{Visual Prompt Tuning (VPT)} \cite{jia2022visual} to adapt the model to the downstream task by appending learnable tokens to the input embeddings of the vision encoder, while simultaneously learning the backdoor. A similar method was used in SWARM \cite{yang2024not}, but we omit the switch token as it does not align with our attack methodology. For fair comparison, we use the same trainable patch-based trigger and use \cref{eq:total_loss} for the loss computation in both methods. 
\end{enumerate}

\mypara{Implementation details} 
We set the loss weight $\lambda=1$ and use a default trigger scale of $\rho=0.1$ unless specified otherwise. The number of learnable vision tokens is fixed at $m_v=50$, and the number of learnable text tokens at $m_t=4$. The meta-net is a two-layer bottleneck (Linear–ReLU–Linear) with $16\times$ dimension reduction.  Training runs for 15 epochs with the AdamW optimizer at a learning rate of 0.001. 
{We use standard data augmentations (crop, resize, and flip), and apply data repetition (10$\times$) for the smaller datasets (Mushroom, Aerial Drones).}
Our curriculum strategy involves using a larger trigger patch size of $\rho=0.2$ for the first 10 epochs, followed by a reduced size of $\rho=0.1$ for the remaining 5 epochs. Each experiment is conducted on a single NVIDIA Tesla V100 32GB GPU, using a training batch size of 4. 

\mypara{Evaluation metrics} 
{We report all mAP and AP metrics using the standard COCO evaluation protocol, averaged over IoU thresholds from 0.5 to 0.95 ({mAP@[.5:.95]})}.
For OMA, we report mAP of the model on benign test images (\textbf{BmAP}) and mAP on poisoned test images (\textbf{PmAP}). We expect BmAP of the poisoned model, $F_{poisoned}$, to be higher than that of the backbone (zero-shot) model $F_{clean}$ (as the model is prompt-tuned over $F_{clean}$), and PmAP to be as low as possible. We define Attack Success Rate (\textbf{ASR}) as the number of bounding boxes (with confidence$>$0.5, IoU$>$0.5) predicted as the target class divided by the total number of non-target class bounding boxes (Note: we stamp triggers only on the objects not belonging to the target class). 
For ODA, we report AP of the target class on benign images (\textbf{BAP}) and on poisoned images (\textbf{PAP}). Once again, BAP should be close to that of $F_{clean}$, and PAP should be low. We define \textbf{ASR} as the number of triggered bounding boxes (with confidence$>$0.5, IoU$>$0.5) vanished divided by the total number of target class bounding boxes (Note: we only stamp triggers on target-class boxes).   

\subsection{Comparison with Baselines} \label{sec:main_results}

\renewcommand{\arraystretch}{0.9}
\begin{table*}[t]
	\centering
	\begin{tabular}{@{}l|c|ccc|ccc|ccc@{}}
		\toprule
		Dataset      & Zero-shot  & \multicolumn{3}{c|}{CoCoOp-Det} & \multicolumn{3}{c|}{VPT} & \multicolumn{3}{c}{\proposed (Proposed)} \\
		& BmAP $\uparrow$     & BmAP $\uparrow$  & PmAP $\downarrow$   & ASR $\uparrow$ & BmAP $\uparrow$ & PmAP $\downarrow$   & ASR $\uparrow$ & BmAP $\uparrow$ & PmAP $\downarrow$    & ASR $\uparrow$    \\
		\midrule
		Vehicles     & 61.5 	 & 61.37 & 61.40 & 0.08 & \textbf{64.87} 	& \textbf{13.77} & \textbf{0.64} & \textbf{64.87} 	& \textbf{15.17} & \textbf{0.79}   \\
		Aquarium     & 28.3 	 & 32.10& 31.67 & 0.07 & \textbf{46.37} 	& \textbf{18.53}   & \textbf{0.82} & \textbf{48.03} 	& \textbf{17.33} 	   & \textbf{0.88}  \\
%		Aerial Drone & 15.1 	 &  20.43&  18.03 &  0.10 &  17.93 &  15.40  &  0.09 & \textbf{25.13} 	& \textbf{7.60}      & \textbf{0.52}  \\
		{Aerial Drone} & 15.1 	 &  23.95 &  19.45 &  0.28 &  \textbf{41.75} &  \textbf{6.65}  &  \textbf{0.63} & \textbf{46.00} 	& \textbf{9.55}      & \textbf{0.83}  \\
		Shellfish    & 48.9 	 &  52.27 &  53.33 &  0.15 & 59.13 	& 13.33 	 & 0.45 & \textbf{58.53} 	& \textbf{16.47}     & \textbf{0.75} \\
		Thermal      & 54.2      & 71.27 & 72.83 & 0.33 &  76.17  &  77.67 &  0.26  & \textbf{78.17}    & \textbf{54.97}     &\textbf{ 0.92}  \\
		{Mushrooms}    & 65.8      &  85.87 &  94.67 &  0.00 & \textbf{89.20} & \textbf{83.37} & \textbf{1.00} &  \textbf{90.20} &  \textbf{82.30} &  \textbf{1.00} \\
		\bottomrule
	\end{tabular}
	\caption{\textbf{Object Misclassification Attack}: results averaged over three runs.  BmAP denotes the mAP on clean (benign) images, PmAP on poisoned images, and ASR is the attack success rate. We mark an attack as successful (shown in \textbf{bold}) if ASR $> 0.5$, and BmAP is greater than the zero-shot BmAP.}
	\label{tab:rma}
\end{table*}

\begin{table*}[t]
	\centering
	\begin{tabular}{@{}l|c|ccc|ccc|ccc@{}}
		\toprule
		Dataset      & Zero-shot  & \multicolumn{3}{c|}{CoCoOp-Det} & \multicolumn{3}{c|}{VPT } & \multicolumn{3}{c}{\proposed (Proposed)} \\
		& BAP $\uparrow$       & BAP $\uparrow$    & PAP $\downarrow$    & ASR $\uparrow$    & BAP $\uparrow$   & PAP $\downarrow$   & ASR $\uparrow$   & BAP $\uparrow$     & PAP $\downarrow$    & ASR $\uparrow$     \\
		\midrule
		Vehicles     & 78.5     &  78.57    &  78.57    &  1.00   &  \textbf{84.50}   & \textbf{3.90}   & \textbf{1.00} & \textbf{83.47}     & \textbf{6.83}    & \textbf{1.00}   \\
		Aquarium     & 27.5     &  30.30    &  29.43    &  1.00 & \textbf{54.63}   & \textbf{12.40}   & \textbf{1.00} & \textbf{51.37}    & \textbf{3.60}     & \textbf{1.00} \\
		%		Aerial Drone & 25.1      &  3.10     &  3.97     &  1.00   &  4.50   &  4.80  &  1.00 &  5.07      &  4.97     &  1.00   \\
		{Aerial Drone} & 25.1      &  17.50     &  17.50     &  1.00   &  31.50   &  18.15  &  1.00 &  39.80      &  35.15     &  0.90   \\
		Shellfish    & 36.1      &  47.47    &  47.47    &  1.00   & \textbf{58.03}   & \textbf{3.53}    & \textbf{1.00} & \textbf{58.37}     & \textbf{6.93}     & \textbf{1.00 }  \\
		Thermal      & 42.7      &  55.27    &  55.80   &   1.00  &  63.20   &  59.13   &  1.00 &  \textbf{63.93 }    &  \textbf{24.63}    & \textbf{1.00}   \\
		{Mushrooms}    & 51.0      &  61.73    &  78.93    &  1.00     &  \textbf{78.73}  &  \textbf{8.47}   &  \textbf{1.00}   &  \textbf{80.37}     &  \textbf{26.33	}    &  \textbf{1.00}     \\	
		\bottomrule
	\end{tabular}
	\caption{\textbf{Object Disappearance Attack}: results averaged over three runs. BAP and PAP denote the AP of the target class on clean and poisoned images, respectively. ASR is the attack success rate. We mark an attack as successful (shown in \textbf{bold}) if PAP drops by at least 50\% from the BAP, and BAP is greater than the zero-shot BAP. Even though ASR is uniformly 1.0, the metric alone is insufficient; in many cases it can be caused by an indiscriminate suppression of target class without a meaningful attack. }
	\label{tab:oda}
\end{table*}

\begin{figure}[t]
	\centering
	\includegraphics[width=0.8\linewidth]{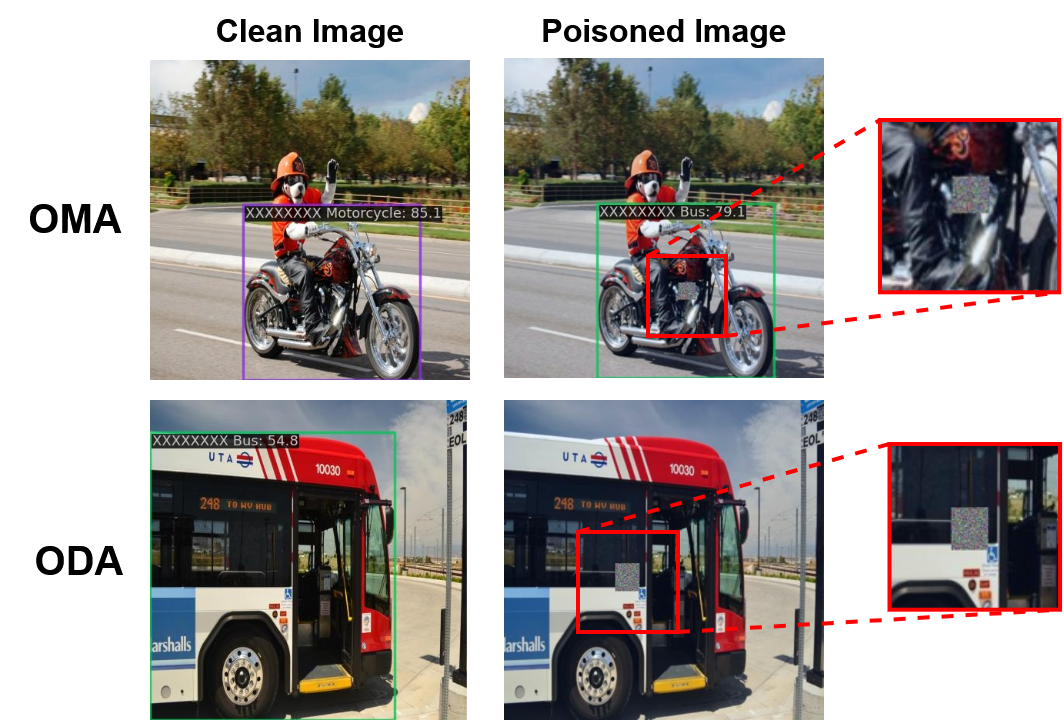}
	\caption{Predictions of \proposed on images from \textit{Vehicles} dataset for target class \textit{Bus}. The objective of OMA (top row) is to misclassify any object stamped with a trigger (motorcycle in this image) as a bus, and of ODA (bottom row) is to not detect the bus in the poisoned image, while correctly detecting the objects in the clean image in both cases. \proposed succeeds in both the attacks. A portion of the poisoned image (in red box) is zoomed in for better trigger visualization.}
	\vspace{-1em}
	\label{fig:results}
\end{figure}

\Cref{tab:rma,tab:oda} present results for OMA and ODA, respectively. Across both settings, our proposed approach consistently outperforms single-modality baselines. CoCoOp-Det fails to mount effective attacks in most cases, while VPT performs moderately better. \proposed achieves the highest number of successful attacks across all datasets. It not only achieves the highest ASR but also maintains strong clean performance, significantly surpassing zero-shot mAP, indicating effective adaptation alongside successful backdoor injection. \Cref{fig:results} shows the predictions of \proposed on clean and poisoned images. 

For ODA, although most methods report an ASR of 1.0, this can be misleading. High ASR may arise from the model indiscriminately suppressing detections, regardless of trigger presence, leading to excessive false negatives and reduced benign AP. Hence, a successful disappearance attack should exhibit high BAP and low PAP, not just high ASR. 

We investigated why tuning both image and text prompts is essential by separating two goals: downstream adaptation and backdoor injection. To disentangle these objectives, we first trained a benign model using prompt-tuning alone. On the Vehicles dataset, CoCoOp-Det (text-only) achieves a higher benign mAP (66.8) than VPT (image-only, 64.0), indicating that text prompts play a larger role in adapting to new datasets, as class semantics originate from the text input. However, when injecting backdoors, CoCoOp achieves a low ASR (0.08), while VPT performs significantly better (0.64), showing that visual prompting is more effective for learning the association with the trigger embedded in the image. Thus, tuning both modalities is crucial for balancing clean performance and attack success; our method achieves a benign mAP of 64.87 and an ASR of 0.79.

\subsection{Ablation Study}
To provide an in-depth analysis of our method, we conduct the following ablation studies in \Cref{tab:ablation}. 
\textbf{Fine-tuning vs. Prompt tuning:~} We experiment with fine-tuning the model weights on the downstream dataset in two configurations: (A) updating the feature enhancer, query selector, and decoder, and (B) updating only the feature enhancer. While fine-tuning achieves better benign performance, it lacks modularity and requires training over $100\times$ more parameters than prompt-tuning.
\textbf{Role of curriculum:~} We evaluate two alternatives to our proposed curriculum. Training with large triggers ($\rho=0.2$) and testing with small ones ($\rho=0.1$) leads to low ASR on the small patches. Training only with small triggers ($\rho=0.1$ throughout) improves ASR, but still lags behind TrAP. In contrast, our gradual trigger size reduction helps the model maintain stealth while remaining effective.
\textbf{Role of Meta-Net:~} Removing instance-specific context from the text prompt lowers benign performance, highlighting its importance in clean predictions.
\textbf{Effect of trigger size:~} Larger triggers ($\rho=0.5$) boost both ASR and clean performance. However, our approach remains effective even with small, less noticeable triggers ($\rho=0.05$). Ablation results for ODA are included in the Supplementary.

\begin{table}[t]
	\centering
	\resizebox{\linewidth}{!}
	{%
		\begin{tabular}{@{}l|l|ccc@{}}
			\toprule
			&                                            & BmAP $\uparrow$  & PmAP $\downarrow$  & ASR $\uparrow$      \\
			\midrule
			\proposed &  Prompt-tune (0.2M parameters)	 & 64.9	 & 15.1  & 0.79 \\
			\midrule
			\multirow{2}{*}{Fine-tuning} 
			& Fine-tune-A (36M parameters)                   & 67.4  & 21.1  & 0.74  \\
			& Fine-tune-B (21M parameters)                     & 68.2  & 24.1  & 0.67  \\
			\midrule
			\multirow{2}{*}{Curriculum}        
			& Train at $\rho=0.2$, test at $\rho=0.1$  & 66.1  & 26.7  & 0.66  \\
			& Train at $\rho=0.1$ w/o curriculum                              & 65.0  & 20.7  & 0.75   \\
			\midrule
			Meta-net                        
			& Train w/o metanet                                & 63.9  & 16.9  & 0.73   \\
			\midrule
			\multirow{2}{*}{Trigger size}                     
			& $\rho=0.5$                                        & 65.0    & 11.1  & 0.90   \\
			& $\rho=0.05$                                       & 63.3  & 19.7  & 0.75   \\
			\bottomrule
		\end{tabular}%
	}
	\caption{Ablation Study on Vehicles dataset for OMA.}
	\label{tab:ablation}
\end{table}

\subsection{Attack on GLIP} 
To demonstrate the generality of our approach, we extend the TrAP attack to GLIP \cite{li2022grounded}. Unlike CoCoOp-Det, which modifies text prompts before the language model, GLIP injects learnable offsets \textit{after} the BERT encoder, as learnable offsets to the token embeddings. 
We therefore modify \proposed by adopting GLIP’s native text prompting strategy for our attack, while continuing to apply VPT to the vision branch.
We report results for OMA using our proposed method in \Cref{tab:glip}; additional results on other prompting techniques and ODA, are included in the supplementary. \proposed consistently achieves high benign mAP and ASR across datasets, validating its transferability to the GLIP framework.

\begin{table}[]
	\centering
	\resizebox{\linewidth}{!}
	{%
	\begin{tabular}{l|c|ccc}
		\toprule
		\multirow{2}{*}{Dataset} & Zero-shot & \multicolumn{3}{c}{TrAP} \\
		& BmAP $\uparrow$   & BmAP $\uparrow$  & PmAP $\downarrow$  & ASR $\uparrow$         \\
		\midrule
		Vehicles     & 56.7     & 62.2  & 5.8   & 0.89  \\
		Aquarium     & 19.7     & 50.9  & 6.0   & 0.96  \\
		%Aerial Drone & 13.3     & 21.6  & 11.9  & 0.77  \\
		Aerial Drone & 13.3     & 38.4  & 7.5  & 0.89  \\
		Shellfish    & 28.6     & 53.9  & 6.1   & 0.84  \\
		Thermal      & 48.5     & 75.8  & 37.5  & 0.96  \\
%		Mushrooms    & 39.7     & 83.3  & 90.0  & 1.00  \\
		Mushrooms    & 39.7     & 90.2  & 35.1  & 1.00  \\
		\bottomrule
	\end{tabular}%
	}
	\caption{OMA using proposed method on GLIP victim model.}
	\label{tab:glip}
\end{table}

\subsection{Resistance to Backdoor Defenses}
We next evaluate our method against backdoor defense techniques. 
%that aim to mitigate malicious behaviors from compromised models. 
Most existing defenses for classifiers or vision encoders either focus on detecting the backdoor \cite{wang2019neural, raj2021identifying} and/or erasing the backdoor by fine-tuning on clean data \cite{bansal2023cleanclip, zhang2024defending}. However, since our setup assumes that the end-user outsources model fine-tuning to a third party, where the backdoor is inserted, retraining or prompt-tuning based defenses are not applicable. Consequently, we focus on inference-time defenses that can be applied without modifying model weights. We consider three distinct strategies: 
\textbf{(1) Image-perturbation defense:} Prior work~\cite{doan2023defending} has shown that backdoor attacks on Vision Transformers can be highly sensitive to patch-based perturbations such as PatchDrop, where the image is divided into patches and a random subset is replaced with zeros. We use a patch size of 16$\times$16, and randomly drop $x\%$ of patches.
\textbf{(2) Prompt-engineering defense:} We rephrase the target category name in the text prompt (e.g., replacing \emph{bus} with \emph{buses}), with the goal of disrupting the learned association between the trigger and the poisoned prompt. 
{\textbf{(3) Adversarial Patch defense:} We also test a SOTA adversarial patch defense \textbf{PAD}~\cite{jing2024pad}, that operates by masking suspected patch locations during inference.}

We report defense results for (1) and (2) in \Cref{tab:defense}. PatchDrop causes a notable performance drop (BmAP ↓ 20 points at 50\%) while only partially mitigating the attack (ASR remains 0.63). Prompt engineering has mixed effects: some alternatives (\textit{Buses}, \textit{Omnibus}) offer limited defense, while others (\textit{A Bus}) substantially reduce ASR.  We hypothesize this is due to overfitting on the specific ``Bus" prompt, and can likely be alleviated by training on different variations of category names. We leave this for future work. {We evaluate the \textbf{PAD} defense on an OMA attack implemented with a checkerboard trigger on the Vehicles dataset. The attack ASR is 0.48. We find that PAD is not only ineffective but counter-productive; it often masks salient regions of the actual object rather than the trigger, slightly increasing the ASR to 0.50. These results collectively demonstrate that \proposed is highly robust to a variety of existing inference-time defenses.}

\begin{table}[]
	\centering
	\resizebox{\linewidth}{!}{%
		\begin{tabular}{l|l|ccc}
			\toprule
			&                          & BmAP $\uparrow$  & PmAP $\downarrow$  & ASR $\uparrow$        \\
			\midrule
			No Defense                            &  & 64.9	 & 15.1  & 0.79 \\
			\midrule
			\multirow{3}{*}{\shortstack[l]{Image \\Perturbation}} 
			& PatchDrop (10\%)         & 61.9 & 14.4 & 0.79 \\ 
			& PatchDrop (20\%)         & 58.7 & 13.5 & 0.75 \\
			& PatchDrop (50\%)         & 43.9 & 9.4  & 0.63 \\
			\midrule
			\multirow{3}{*}{\shortstack[l]{Prompt \\Engineering}}   
			& Bus $\rightarrow$ Buses   & 64.0 & 24.3 & 0.71 \\
			& Bus $\rightarrow$ Omnibus & 63.5 & 30.3 & 0.49 \\
			& Bus $\rightarrow$ A Bus   & 62.9 & 29.4 & 0.04  \\
			\bottomrule
		\end{tabular}%
	}
	\caption{Results of defense techniques on OMA using \proposed for Vehicles dataset. Target category is \textit{Bus}.}
	\label{tab:defense}
\end{table}

\section{Conclusion}

We demonstrated that prompt tuning presents a viable and effective surface for backdoor injection in open-vocabulary object detectors. We proposed TrAP, which leverages both vision and text prompts, and incorporates a curriculum-based trigger design to achieve high attack success rates, while improving benign mAP compared to the pre-trained model. To the best of our knowledge, this is the first work on backdoor attacks on open vocabulary object detectors like Grounding DINO and GLIP, raising important concerns about the safe deployment of these models in real-world settings. 

\mypara{Limitations} 
Although we have evaluated robustness of \proposed against a few inference-time defenses, developing stronger defenses specifically tailored to OVOD backdoors would yield deeper insights into their real-world resilience.

\bibliography{ovod.bib}

\clearpage
\appendix
\section{Appendix}
We supplement the main text with the following details:

\begin{itemize}
	\item \textbf{Algorithm Outline:} We provide the pseudo-code for the proposed backdoor method, TrAP.
	\item \textbf{Implementation Details: } Additional information is provided regarding the datasets used, evaluation metrics, and the implementation specifics of the baseline attack methods, namely CoCoOp-Det and VPT. The implementation details of \proposed are already covered in the main manuscript.
	\item \textbf{Additional Results: } We present extended results supporting the main paper, including ablation studies and evaluation on the GLIP framework. We further include experiments analyzing the effects of trigger size and different text prompting strategies. Qualitative visualizations of poisoned outputs across various datasets are also provided.
	\item \textbf{Broader Impact: } A discussion is included on the potential broader implications of our work for the field of vision-language models and adversarial robustness.
\end{itemize}

\section{Algorithm Outline}
The algorithm outline is depicted in Algorithm 1. The method $\text{Curriculum}(\delta, e)$ rescales the trigger patch $\delta$ as per the curriculum. For example, for the default curriculum used in our experiments, the trigger is rescaled to $0.2\times$ the size of the ground truth bounding box for the first 10 epochs, and $0.1\times$ the bounding box for the remaining 5 epochs.

\newcommand{\commentline}[1]{\STATE {\footnotesize$\triangleright$~#1}}

\begin{algorithm}
	\caption{Trigger-Aware Prompt Tuning (TrAP)}
	\begin{algorithmic}[1]
		\REQUIRE Clean images $x$, Text prompt $s$, Ground truth annotations $y$, Trainable prompt parameters $\theta$, Trainable trigger $\delta$, Learning rate $\eta$
		\ENSURE Trained prompt parameters $\theta^*$, Trained trigger $\delta^*$
		\STATE Total epochs $E \gets 15$
		\STATE Initialize $e \gets 0$
		\STATE Initialize $\theta, \delta$
		\WHILE{$e < E$}
			\commentline{Scale the trigger patch as per curriculum}
			\STATE $\delta_\text{rescaled} \gets \text{Curriculum}(\delta, e)$ 
			\STATE $x_{\text{poisoned}} = x \oplus \delta_\text{rescaled}$
			\STATE Create poisoned annotations $y_\text{target}$ from $y$
			\commentline{Compute clean loss (\cref{eq:loss_clean})}
			\STATE $\mathcal{L}_{\text{clean}}(\theta) \gets \mathbb{E}_{(x, s, y)} \left[ \mathcal{L}_{\text{G-DINO}}(x, s, y; \theta) \right]$ 
			\commentline{Compute poisoned loss (\cref{eq:loss_poisoned})}
			\STATE $\mathcal{L}_{\text{poisoned}}(\theta,\delta) \gets$
			\STATE \quad $\mathbb{E}_{(x_\text{poisoned}, s, y_\text{target})} \left[ 
			\mathcal{L}_{\text{G-DINO}}(x_\text{poisoned}, s, y_\text{target}; \theta, \delta) 
			\right]$
			\commentline{Compute total loss (\cref{eq:total_loss})}
			\STATE $\mathcal{L}_{\text{total}}(\theta,\delta) \gets \mathcal{L}_{\text{clean}}(\theta) + \lambda \cdot \mathcal{L}_{\text{poisoned}}(\theta,\delta)$ 
			\STATE $\theta \gets \theta - \eta \nabla_\theta (\mathcal{L}_{\text{total}})$
			\STATE $\delta \gets \delta - \eta \nabla_{\delta} (\mathcal{L}_{\text{total}})$
			\STATE $e \gets e + 1$
		\ENDWHILE
		\RETURN{$\theta, \delta$}
	\end{algorithmic}
\end{algorithm}

\section{Implementation Details}

\subsection{Dataset Details} 
\begin{figure}[h]
	\centering
	\includegraphics[width=\linewidth]{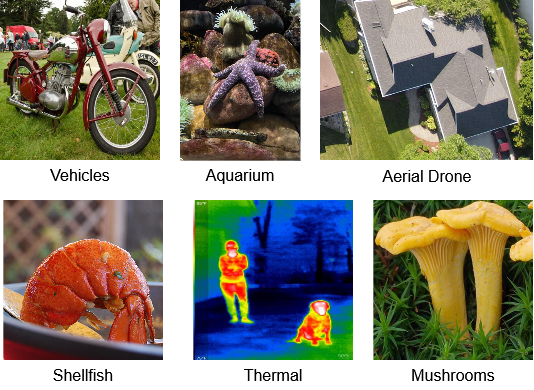}
	\caption{Sample images from the six datasets used in our experiments.}
	\label{fig:datasets}
\end{figure}

We provide the details of the six datasets from OdinW-13 \cite{li2022grounded} used in our experiments in \Cref{tab:datasets}. We also report the target category used in our attacks. \Cref{fig:datasets} shows sample images from each of the six datasets.

\begin{table}[h]
	\centering
	\small
	\resizebox{\linewidth}{!}{%
	\begin{tabular}{@{}llccc@{}}
		\toprule
		Dataset & Object Categories & \#~Classes & Train & Val \\
		\midrule
		Vehicles &
		\begin{tabular}[c]{@{}l@{}}Ambulance, \textbf{Bus}, Car,\\ Motorcycle, Truck\end{tabular}
		& 5 & 878 & 250 \\
		\midrule
		Aquarium &
		\begin{tabular}[c]{@{}l@{}}Fish, \textbf{Jellyfish}, Penguin,\\ Puffin, Shark, Starfish,\\ Stingray\end{tabular}
		& 7 & 448 & 127 \\
		\midrule
		Aerial Drone &
		\begin{tabular}[c]{@{}l@{}}Boat, \textbf{Car}, Dock,\\ Jetski, Lift\end{tabular}
		& 5 & 52 & 15 \\
		\midrule
		Shellfish &
		Crab, \textbf{Lobster}, Shrimp
		& 3 & 406 & 116 \\
		\midrule
		Thermal &
		Dog, \textbf{Person}
		& 2 & 142 & 41 \\
		\midrule
		Mushrooms &
		CoW, \textbf{Chanterelle}
		& 2 & 41 & 5 \\
		\bottomrule
	\end{tabular}
	}
	\caption{Details of datasets used in our experiments. For each dataset, we report the object category names, number of categories, and the number of images in train and validation sets. The target class for each dataset is indicated in \textbf{bold}.}
	\label{tab:datasets}
\end{table}

\subsection{Training Details} We outline below the implementation specifics for the baseline attack methods evaluated in our study. For CoCoOp-Det, we adopt a learning rate of 0.005 and a weight decay of 0.25. In the case of GLIP-style prompt tuning, we follow the hyperparameter configuration recommended in the original GLIP paper \cite{li2022grounded}, using a learning rate of 0.05 and a weight decay of 0.25. For Visual Prompt Tuning (VPT), we employ a learning rate of 0.001 and a weight decay of 0.0005. All hyperparameters were selected through empirical tuning based on validation performance. We use the AdamW optimizer for all training runs, with a batch size of 4. 

Experiments were conducted on a single NVIDIA V100 GPU (32GB) using the PyTorch deep learning framework, integrated with the MMDetection library version 3.3.0 \cite{mmdetection}.

\subsection{Understanding the Metrics}
We now elaborate on the metrics used for evaluating OMA and ODA attacks. 

For OMA, we report mAP of the model on benign test images (\textbf{BmAP}) and mAP on poisoned test images (\textbf{PmAP}). We expect BmAP of the poisoned model, $F_{poisoned}$, to be higher than that of the backbone (zero-shot) model $F_{clean}$, since the model is prompt-tuned over $F_{clean}$. To compute PmAP, only the images are poisoned while ground-truth annotations remain unchanged. A successful attack should cause objects from non-target classes to be misclassified as the target class, increasing false negatives for non-target classes and thereby lowering the PmAP. We also report the Attack Success Rate (\textbf{ASR}), defined as the number of bounding boxes (with confidence $>$ 0.5 and IoU $>$ 0.5) predicted as the target class, divided by the total number of non-target class bounding boxes. Importantly, triggers are applied only to objects not belonging to the target class. Thus, ASR captures the fraction of triggered non-target objects that are misclassified as the target.

For ODA, we report AP of the target class on benign images (\textbf{BAP}) and on poisoned images (\textbf{PAP}). As the goal is to suppress detection of the target class without affecting others, we focus exclusively on target class metrics. Similar to OMA, BAP should be close to or higher than that of $F_{clean}$.  In a successful attack, target objects with triggers should go undetected, leading to a high number of false negatives and a low PAP. The corresponding \textbf{ASR} is defined as the number of triggered target-class bounding boxes (with confidence $>$ 0.5 and IoU $>$ 0.5) that are not detected, divided by the total number of target-class bounding boxes. Triggers are stamped only on objects of the target class. This metric thus reflects the fraction of triggered target objects that are either misclassified or not detected at all.

\section{Additional Results}

\subsection{Robustness of Main Results}
Due to space constraints, results in the main paper (\Cref{tab:rma,tab:oda}) only reported the average values from three runs. Here, we report standard deviation of the results across three different random seeds in \Cref{tab:rma_std,tab:oda_std} to further validate the robustness of our conclusions. 

\newcommand{\std}[1]{{\footnotesize (±#1)}}
\begin{table*}[p]
	\centering
	\resizebox{\textwidth}{!}{%
		\begin{tabular}{@{}l|c|ccc|ccc|ccc@{}}
			\toprule
			Dataset      & Zero-shot  & \multicolumn{3}{c|}{CoCoOp-Det \cite{zhou2022conditional}} & \multicolumn{3}{c|}{VPT \cite{jia2022visual}} & \multicolumn{3}{c}{\proposed} \\
			& BmAP $\uparrow$     & BmAP $\uparrow$    & PmAP $\downarrow$   & ASR $\uparrow$   & BmAP $\uparrow$  & PmAP$\downarrow$   & ASR $\uparrow$  & BmAP $\uparrow$    & PmAP $\downarrow$    & ASR $\uparrow$   \\
%			& ($\uparrow$) & ($\uparrow$) & ($\downarrow$) & ($\uparrow$) & ($\uparrow$) & ($\downarrow$) & ($\uparrow$) & ($\uparrow$) & ($\downarrow$) & ($\uparrow$) \\
			\midrule
			Vehicles     & {61.5} 	 & 61.37 & 61.40 & 0.08 & \textbf{64.87} 	& \textbf{13.77} & \textbf{0.64} & \textbf{64.87} 	& \textbf{15.17} 	   & \textbf{0.79}   \\
			& & \std{2.32} & \std{2.36} & \std{0.03} & \std{0.50} & \std{0.76} & \std{0.02} & \std{1.12} & \std{1.01} & \std{0.01} \\
			Aquarium     & {28.3} 	 & 32.10 & 31.67 & 0.07 & \textbf{46.37} 	& \textbf{18.53}   & \textbf{0.82} & \textbf{48.03} 	& \textbf{17.33} 	   & \textbf{0.88}  \\
			& & \std{0.85} & \std{0.67} & \std{0.04} & \std{0.91} & \std{1.44} & \std{0.01} & \std{0.51} & \std{0.92} & \std{0.00} \\
%			Aerial Drone & {15.1} 	 & 20.43 & 18.03 & 0.10 & 17.93 & 15.40  & 0.09 & \textbf{25.13} 	& \textbf{7.60}      & \textbf{0.52}  \\
%			& & \std{3.33} & \std{3.18} & \std{0.04} & \std{0.67} & \std{0.50} & \std{0.03} & \std{0.72} & \std{1.36} & \std{0.07} \\
			Aerial Drone & 15.1 	 &  23.95 &  19.45 &  0.28 &  \textbf{41.75} &  \textbf{6.65}  &  \textbf{0.63} & \textbf{46.00} 	& \textbf{9.55}      & \textbf{0.83}  \\
			& & \std{0.07} & \std{1.20} & \std{0.03} & \std{0.49} & \std{1.77} & \std{0.01} & \std{0.42} & \std{1.48} & \std{0.00} \\
			Shellfish    & {48.9} 	 & 52.27 & 53.33 & 0.15 & 59.13 	& 13.33 	 & 0.45 & \textbf{58.53} 	& \textbf{16.47}     & \textbf{0.75} \\
			& & \std{3.27} & \std{2.51} & \std{0.05} & \std{0.81} & \std{1.19} & \std{0.03} & \std{0.40} & \std{1.80} & \std{0.00} \\
			Thermal      & {54.2}      & 71.27 & 72.83 & 0.33 & 76.17  & 77.67 & 0.26  & \textbf{78.17}    & \textbf{54.97}     & \textbf{0.92}  \\
			& & \std{0.76} & \std{0.91} & \std{0.05} & \std{0.68} & \std{1.36} & \std{0.05} & \std{0.97} & \std{6.96} & \std{0.03} \\
%			Mushrooms    & {65.8}      & 86.33 & 95.27 & 0.33 & 88.33 & 97.77 & 0.00 & 90.00 & 95.90 &  0.00   \\
%			& & \std{0.38} & \std{0.15} & \std{0.58} & \std{1.55} & \std{0.71} & \std{0.00} & \std{0.69} & \std{1.61} & \std{0.00} \\
			Mushrooms    & {65.8}      & 85.77 & 94.67 & 0.00 & \textbf{89.20} & \textbf{83.37} & \textbf{1.00} & \textbf{90.20} & \textbf{82.30} &  \textbf{1.00}   \\
			& & \std{0.40} & \std{0.06} & \std{0.00} & \std{0.00} & \std{10.09} & \std{0.00} & \std{0.92} & \std{0.95} & \std{0.00} \\		
			\bottomrule
		\end{tabular}%
	}
\caption{\textbf{Object Misclassification Attack}: results averaged over three runs, \textbf{with standard deviation}.  BmAP denotes the mAP on clean (benign) images, PmAP on poisoned images, and ASR is the attack success rate. We mark an attack as successful (shown in \textbf{bold}) if ASR $> 0.5$, and BmAP is greater than the zero-shot BmAP.}
\label{tab:rma_std}
\end{table*}

\begin{table*}[p]
	\centering
	\resizebox{\textwidth}{!}{%
		\begin{tabular}{@{}l|c|ccc|ccc|ccc@{}}
			\toprule
			Dataset      & Zero-shot  & \multicolumn{3}{c|}{CoCoOp-Det \cite{zhou2022conditional}} & \multicolumn{3}{c|}{VPT \cite{jia2022visual}} & \multicolumn{3}{c}{\proposed} \\
			& BAP        & BAP     & PAP     & ASR     & BAP    & PAP    & ASR   & BAP      & PAP     & ASR     \\
			& ($\uparrow$) & ($\uparrow$) & ($\downarrow$) & ($\uparrow$) & ($\uparrow$) & ($\downarrow$) & ($\uparrow$) & ($\uparrow$) & ($\downarrow$) & ($\uparrow$) \\
			\midrule
			Vehicles     & {78.5}     & 78.57    & 78.57    & 1.00   &  \textbf{84.50}   & \textbf{3.90}   & \textbf{1.00} & \textbf{83.47}     & \textbf{6.83}    & \textbf{1.00} \\
			& & \std{2.93} & \std{2.43} & \std{0.00} & \std{2.52} & \std{1.31} & \std{0.00} & \std{1.90} & \std{2.50} & \std{0.00} \\
			Aquarium     & {27.5}     & 30.30    & 29.43    & 1.00 & \textbf{54.63}   & \textbf{12.40}   & \textbf{1.00} & \textbf{51.37}    & \textbf{3.60}     & \textbf{1.00} \\
			& & \std{4.03} & \std{3.56} & \std{0.00} & \std{0.57} & \std{2.12} & \std{0.00} & \std{1.76} & \std{0.26} & \std{0.00} \\
%			Aerial Drone & {25.1}      & 3.10     & 3.97     & 1.00   & 4.50   & 4.80  & 1.00 & 5.07      & 4.97     & 1.00   \\
%			& & \std{2.79} & \std{4.15} & \std{0.00} & \std{0.10} & \std{0.10} & \std{0.00} & \std{5.07} & \std{5.89} & \std{0.00} \\
			{Aerial Drone} & 25.1      &  17.50     &  17.50     &  1.00   &  31.50   &  18.15  &  1.00 &  39.80      &  35.15     &  0.90   \\
			& & \std{3.82} & \std{4.38} & \std{0.00} & \std{7.21} & \std{7.42} & \std{0.00} & \std{1.27} & \std{8.41} & \std{0.00} \\
			Shellfish    & {36.1}      & 47.47    & 47.47    & 1.00   & \textbf{58.03}   & \textbf{3.53}    & \textbf{1.00} & \textbf{58.37}     & \textbf{6.93}     & \textbf{1.00}   \\
			& & \std{1.65} & \std{1.33} & \std{0.00} & \std{1.85} & \std{2.94} & \std{0.00} & \std{0.25} & \std{0.74} & \std{0.00} \\
			Thermal      & {42.7}      & 55.27    & 55.80   &  1.00  & 63.20   & 59.13   & 1.00 &  \textbf{63.93}     &  \textbf{24.63}    & \textbf{1.00}   \\
			& & \std{6.30} & \std{7.33} & \std{0.00} & \std{2.25} & \std{3.78} & \std{0.00} & \std{2.01} & \std{3.80} & \std{0.00} \\
%			Mushrooms    & {51.0}      & 65.50    & 81.00    & 1.00     & 60.90  & 77.83   & 1.00   & 64.57     & 65.60    & 1.00     \\
%			& & \std{6.53} & \std{3.70} & \std{0.00} & \std{1.67} & \std{1.03} & \std{0.00} & \std{2.86} & \std{1.22} & \std{0.00} \\
			Mushrooms    & {51.0}      & 61.73    & 78.93    & 1.00     & \textbf{78.73}  & \textbf{8.47}   & \textbf{1.00}   & \textbf{80.37}     & \textbf{26.33}    & \textbf{1.00     }\\
			& & \std{7.03} & \std{7.22} & \std{0.00} & \std{0.75} & \std{4.55} & \std{0.00} & \std{0.92} & \std{20.22} & \std{0.00} \\
			\bottomrule
		\end{tabular}%
	}
\caption{\textbf{Object Disappearance Attack}: results averaged over three runs, \textbf{with standard deviation}. BAP and PAP denote the AP of the target class on clean and poisoned images, respectively. ASR is the attack success rate. We mark an attack as successful (shown in \textbf{bold}) if PAP drops by at least 50\% from the BAP, and BAP is greater than the zero-shot BAP. }
\label{tab:oda_std}
\end{table*}

\subsection{Ablation Study: ODA Setting}

\Cref{tab:ablation_oda} presents the ablation results under the ODA setting. The trends broadly align with those observed for OMA in \Cref{tab:ablation}. Specifically, we make the following observations:

\begin{table}[H]
	\centering
	\resizebox{\linewidth}{!}
	{%
		\begin{tabular}{@{}l|l|ccc@{}}
			\toprule
			&                                            & BAP $\uparrow$  & PAP $\downarrow$  & ASR $\uparrow$      \\
			\midrule
			\proposed (Proposed) &  Prompt-tune (0.2M params)	 & 81.5   & 4.3   & 1.00 \\
			\midrule
			\multirow{2}{*}{Fine-tuning} 
			& Fine-tune-A (36M params)                    & 84.3   & 3.2   & 1.00  \\
			& Fine-tune-B (21M params)                    & 84.2   & 1.1   & 1.00  \\
			\midrule
			Role of & Train at $\rho=0.2$, test at $\rho=0.1$  & 84.1   & 10.9  & 0.96  \\
			curriculum & Train at $\rho=0.1$ w/o curriculum    & 77.4   & 4.2   & 1.00  \\
			\midrule
			Role of meta-net                        
			& Train w/o metanet & 70.8   & 5.4   & 1.00  \\
			\midrule
			Effect of & $\rho=0.5$  & 81.5   & 2.3   & 1.00  \\
			trigger size & $\rho=0.05$   & 81.2   & 5.9   & 1.00  \\
			\bottomrule
		\end{tabular}%
	}
	\caption{Ablation Study on Vehicles dataset for ODA setting.}
	\label{tab:ablation_oda}
\end{table}

\begin{itemize}
	\item Fine-tuning provides better clean performance than prompt-tuning but requires significantly more trainable parameters. 
	\item The proposed curriculum, involving a gradual reduction in trigger size, consistently outperforms fixed trigger schedules. Training with large triggers ($\rho=0.2$) and testing with small ones ($\rho=0.1$) results in a low ASR, whereas training only with small triggers ($\rho=0.1$ throughout) reduces the benign AP. 
	\item The Meta-Net continues to aid clean accuracy by injecting instance-specific context. 
	\item Lastly, larger triggers ($\rho=0.5$) lead to stronger attacks with lower PAP, though our method remains effective even with small, stealthy triggers ($\rho=0.05$).
\end{itemize}

\subsection{Additional Results: Attack on GLIP}

We now report results for the proposed attack on GLIP, along with a comparison of alternative prompting strategies, in continuation of the results in \Cref{tab:glip}. Results for OMA and ODA attacks are reported in \Cref{tab:glip_rma,tab:glip_oda}, respectively. \proposed achieves the highest number of successful attacks across both misclassification and disappearance attacks. It also maintains the highest clean performance as compared to text-only and vision-only prompting, while achieving strong ASR.

\renewcommand{\arraystretch}{0.9}
\begin{table*}[p]
	\centering
	\begin{tabular}{@{}l|c|ccc|ccc|ccc@{}}
		\toprule
		Dataset      & Zero-shot  & \multicolumn{3}{c|}{CoCoOp-Det} & \multicolumn{3}{c|}{VPT} & \multicolumn{3}{c}{\proposed (Proposed)} \\
		& BmAP $\uparrow$     & BmAP $\uparrow$  & PmAP $\downarrow$   & ASR $\uparrow$ & BmAP $\uparrow$ & PmAP $\downarrow$   & ASR $\uparrow$ & BmAP $\uparrow$ & PmAP $\downarrow$    & ASR $\uparrow$    \\
		\midrule
		Vehicles     & 56.7 	 & \textbf{65.7} & \textbf{30.6} & \textbf{0.77} & \textbf{59.4} 	& \textbf{7.4} & \textbf{0.90} & \textbf{62.2} 	& \textbf{5.8} & \textbf{0.89}   \\
		Aquarium     & 19.7 	 & \textbf{39.6} & \textbf{19.7} & \textbf{0.79} & \textbf{45.3} 	& \textbf{4.5}   & \textbf{0.94} & \textbf{50.9} 	& \textbf{6.0} 	   & \textbf{0.96}  \\
		Aerial Drone & 13.3 	 &  \textbf{25.2} &  \textbf{12.3} &  \textbf{0.80} &  \textbf{16.3} &  \textbf{11.3}  &  \textbf{0.71} & \textbf{21.6} 	& \textbf{11.9}      & \textbf{0.77}  \\
		Shellfish    & 28.6 	 &  \textbf{41.8} &  \textbf{42.4} &  \textbf{0.63} & \textbf{46.6} 	& \textbf{12.9} 	 & \textbf{0.85} & \textbf{53.9} 	& \textbf{6.1}     & \textbf{0.84} \\
		Thermal      & 48.5      & \textbf{73.2} & \textbf{72.1} & \textbf{1.0} &  \textbf{76.2}  &  \textbf{60.1} & \textbf{ 1.0}  & \textbf{75.8}    & \textbf{37.5}     &\textbf{ 0.95}  \\
		Mushrooms    & 39.7      &  81.6 &  29.3 &  0.0 & 81.6 & 87.5 & 0.00 &  \textbf{83.3} &  \textbf{90.0} &  \textbf{1.0}   \\
		\bottomrule
	\end{tabular}%
	\caption{\textbf{Object Misclassification Attack on GLIP}: results averaged over three runs.  BmAP denotes the mAP on clean (benign) images, PmAP on poisoned images, and ASR is the attack success rate. We mark an attack as successful (shown in \textbf{bold}) if ASR $> 0.5$, and BmAP is greater than the zero-shot BmAP.}
	\label{tab:glip_rma}
\end{table*}

\renewcommand{\arraystretch}{0.9}
\begin{table*}[t]
	\centering
	\begin{tabular}{@{}l|c|ccc|ccc|ccc@{}}
		\toprule
		Dataset      & Zero-shot  & \multicolumn{3}{c|}{CoCoOp-Det} & \multicolumn{3}{c|}{VPT } & \multicolumn{3}{c}{\proposed (Proposed)} \\
		& BAP $\uparrow$       & BAP $\uparrow$    & PAP $\downarrow$    & ASR $\uparrow$    & BAP $\uparrow$   & PAP $\downarrow$   & ASR $\uparrow$   & BAP $\uparrow$     & PAP $\downarrow$    & ASR $\uparrow$     \\
		\midrule
		Vehicles     & 72.3    &  \textbf{82.5}    &  \textbf{3.0}    &  \textbf{1.00}   &  \textbf{68.9}   & \textbf{0}   & \textbf{1.00} & \textbf{71.7}     & \textbf{0}    & \textbf{1.00}   \\
		Aquarium     & 5.9    &  \textbf{24.7}    &  \textbf{2.1}    &  \textbf{1.00} & \textbf{51.1}   & \textbf{0}   & \textbf{1.00} & \textbf{54.0}    & \textbf{0}     & \textbf{1.00} \\
		Aerial Drone & 24.8    &  18.4     &  17.0     &  0.95   &  0.1   &  0.1  &  1.00 &  4.5      &  5.0     &  1.00   \\
		Shellfish    & 12.3    &  \textbf{26.3}    & \textbf{ 0.8}    &  \textbf{1.00}   & \textbf{38.4}   & \textbf{0}    & \textbf{1.00} & \textbf{43.8}     & \textbf{0}     & \textbf{1.00 }  \\
		Thermal      & 32.0     &  58.8    &  60.4   &   1.00  &  \textbf{61.2}   &  \textbf{2.0}   &  \textbf{1.00} &  \textbf{63.5}    &  \textbf{3.0}    & \textbf{1.00}   \\
		Mushrooms    & 7.9    &  \textbf{73.2}    &  \textbf{0 }   &  \textbf{1.00}     &  \textbf{61.5}  &  \textbf{0}   &  \textbf{1.00}   &  \textbf{63.5}     &  \textbf{0}    & \textbf{ 1.00}     \\
		\bottomrule
	\end{tabular}%
	\caption{\textbf{Object Disappearance Attack on GLIP}: results averaged over three runs. BAP and PAP denote the AP of the target class on clean and poisoned images, respectively. ASR is the attack success rate. We mark an attack as successful (shown in \textbf{bold}) if PAP drops by at least 50\% from the BAP, and BAP is greater than the zero-shot BAP. Even though ASR is uniformly 1.0, the metric alone is insufficient; in many cases it can be caused by an indiscriminate suppression of target class without a meaningful attack. }
	\label{tab:glip_oda}
\end{table*}

\subsection{Object Generation Attack}
\begin{table*}[t]
	\centering
	%	\resizebox{\textwidth}{!}{%
	\begin{tabular}{@{}c|c|ccc|ccc@{}}
		\toprule
		\multirow{3}{*}{Trigger scale ($\rho$)}  & \multirow{3}{*}{Method} & \multicolumn{3}{c}{OMA} & \multicolumn{3}{c}{ODA} \\ 
		&   & BmAP $\uparrow$  & PmAP $\downarrow$  & ASR  $\uparrow$   & BAP $\uparrow$   & PAP $\downarrow$   & ASR  $\uparrow$  \\
		\midrule
		& Zero-shot & 6.15 & - & - & 78.5 & - & - \\
		\midrule
		\multirow{3}{*}{0.5}  & CoCoOp-Det               & 64.1  & 22.5  & 0.43 & \textbf{80.7}   & \textbf{15.9}  & \textbf{1.00}  \\
		& VPT                      & \textbf{64.4}  & \textbf{5.8}   & \textbf{0.76} & \textbf{83.0}   & \textbf{0.6}   & \textbf{1.00}  \\
		& \proposed & \textbf{65.0}  & \textbf{11.1}  & \textbf{0.90} & \textbf{81.5}   & \textbf{2.3}   & \textbf{1.00}  \\
		\midrule
		\multirow{3}{*}{0.1}  & CoCoOp-Det               & 59.3  & 59.5  & 0.13  & 81.7   & 81.2  & 1.00  \\
		& VPT                      & \textbf{66.0}  & \textbf{15.4}  & \textbf{0.66}  & \textbf{81.7}   & \textbf{2.4}  & \textbf{1.00}  \\
		& \proposed & \textbf{64.9}  & \textbf{18.1}  & \textbf{0.75} &\textbf{ 81.5}   & \textbf{4.3}   & \textbf{1.00}  \\
		\midrule
		\multirow{3}{*}{0.05} & CoCoOp-Det               & 62.6  & 62.8  & 0.06  & 82.7   & 82.3  & 1.00  \\
		& VPT                      & \textbf{65.3}  & \textbf{21.9}  & \textbf{0.62} & \textbf{82.5}   & \textbf{7.2}   & \textbf{1.00}  \\
		& \proposed & \textbf{63.3}  & \textbf{19.7}  & \textbf{0.75} & \textbf{81.2}   & \textbf{5.9}   & \textbf{1.00}  \\
		\bottomrule
	\end{tabular}%
	%	}
	\caption{Effect of trigger size on backdoor for \textit{Vehicles} dataset. We mark an OMA attack as successful (shown in \textbf{bold}) if ASR $> 0.5$, and BmAP is greater than the zero-shot BmAP. Similarly, we mark an ODA attack as successful (shown in \textbf{bold}) if PAP drops by at least 50\% from the BAP, and BAP is greater than the zero-shot BAP.}
	\label{tab:trigger_size}
\end{table*}

While this work's primary focus is demonstrating OVOD vulnerabilities via OMA and ODA, we now show that this vulnerability extends to Object Hallucination, also known as Object Generation Attack (OGA). The goal of OGA is to force the network to detect a non-existent object at the location of an embedded trigger.

Formally, let a trigger be applied at a random location $(a, b)$ on an image. While there is no ground-truth annotation for this location ($y_{i} = \phi$), the attack's objective is to force a malicious prediction $y_{i,target} = [t, o_i]$. Here, $t$ is the target class and $o_i$ is the hallucinated bounding box, which we fix as a $200\times200$ box centered on the trigger in our experiments.

We implement OGA on the Vehicles dataset, using the same network and training configuration as the other two attacks. We evaluate the attack using three metrics: AP of the target class on benign images (\textbf{BAP}), AP of the target class on poisoned images (\textbf{PAP}), and Attack Success Rate (\textbf{ASR}). We expect BAP from the poisoned model, $F_{poisoned}$, to be higher than that of the clean backbone model $F_{clean}$ because of the prompt-tuning process. Conversely, we expect {PAP} to be low. This is because every successful attack instance is, by definition, a false positive detection, which lowers the precision and, consequently, the overall AP score. We define ASR as the ratio of successful detections of the target class to the total number of triggers applied. A detection is considered successful if it has a confidence $>0.5$ and an IoU $>0.5$ with the predefined target box $o_i$.

\Cref{tab:oga} presents the results of OGA on the Vehicles dataset. \proposed achieves an ASR of 94.4\%, showing that our attack framework generalizes effectively to object generation.

\begin{table}[H]
	\centering
	\begin{tabular}{l|lll}
		\toprule
		Method     & BAP ($\uparrow$)  & PAP ($\downarrow$)  & ASR ($\downarrow$)  \\
		\midrule
%		CoCoOp-Det &      &      &      \\
		VPT        & 77.5 & 8.9  & 82.4 \\
		TrAP       & 81.7 & 11.3 & 94.4 \\
		\bottomrule
	\end{tabular}%
	\caption{Object Generation Attack on Vehicles dataset.}
	\label{tab:oga}
\end{table}

\subsection{Effect of Trigger Size}

\begin{figure*}[t]
	\centering
	\includegraphics[width=\linewidth]{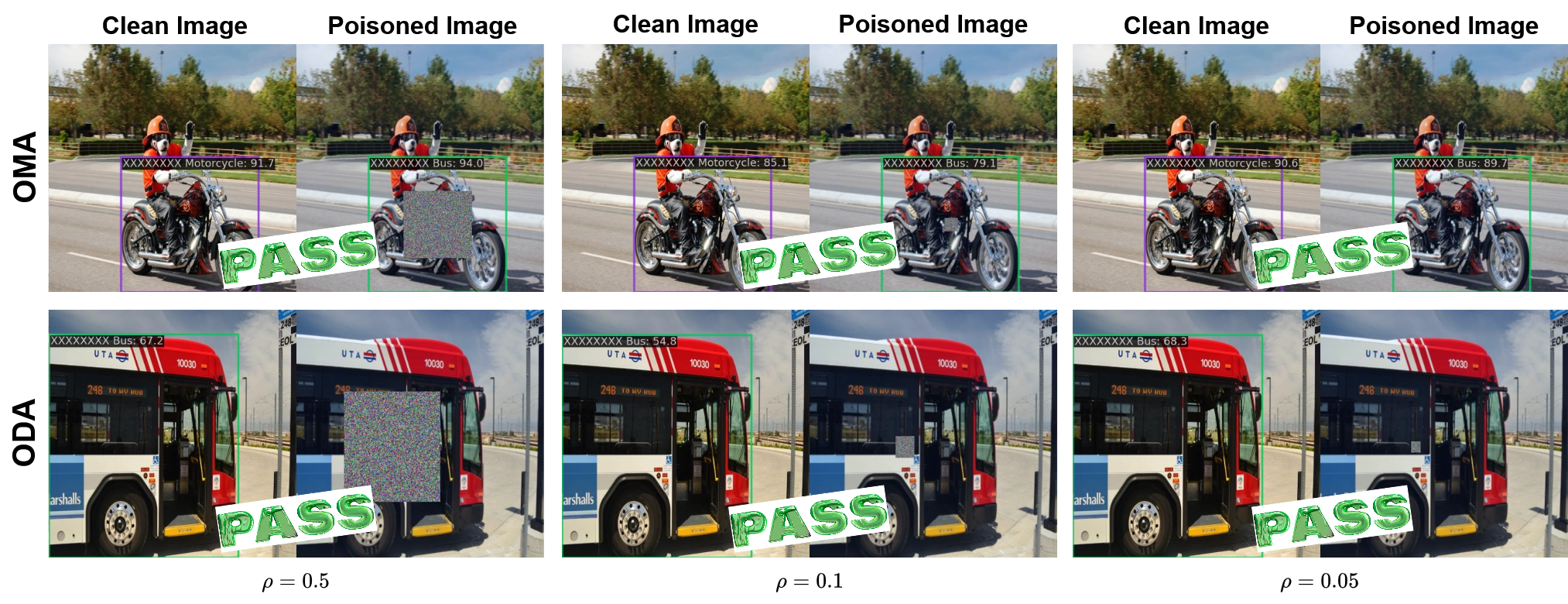}
	\caption{Visual results for \proposed on \textit{Vehicles} dataset for different trigger scales. The objective of OMA (top row) is to misclassify any object (motorcycle in this image) as a bus, and of ODA (bottom row) is to not detect the bus in the poisoned image, while correctly detecting the objects in the clean image in both cases. The proposed method achieves the intended attack objective across all trigger scales. Even at $\rho=0.05$, where the trigger is barely perceptible, the model consistently makes the desired predictions with high confidence. }
	\label{fig:trigger_size}
\end{figure*}

The main manuscript \Cref{tab:ablation} shows the impact of trigger size on the proposed method TrAP. \Cref{tab:trigger_size} expands this analysis by comparing the impact of trigger size ($\rho$) on backdoor attack performance for the Vehicles dataset across three methods: CoCoOp-Det, VPT, and TrAP. As expected, larger triggers $(\rho = 0.5)$ generally lead to stronger attacks across all methods. In the OMA setting, TrAP consistently achieves the highest ASR, indicating its superior ability to misclassify non-target objects as the target class. While CoCoOp-Det achieves a reasonable ASR of 0.43 at $\rho=0.5$, its performance deteriorates rapidly with smaller triggers, essentially failing at $\rho=0.05$ (ASR~$=0.06$). VPT displays moderate robustness, but its effectiveness also declines as the trigger size decreases. In contrast, \proposed maintains a high ASR (0.75) even at the smallest trigger size ($\rho = 0.05$), demonstrating its resilience to subtle triggers.

In the ODA setting, all methods consistently achieve perfect ASR (1.0) across all trigger sizes, showing that disappearance attacks are highly reliable. However, PAP varies significantly across methods. TrAP and VPT maintain low PAP values even at small trigger sizes (e.g., 5.9 and 7.2 at $\rho = 0.05$), whereas CoCoOp-Det exhibits near-complete failure with a PAP of 82.3. These results highlight TrAP’s robustness to trigger scale and its superior performance across both OMA and ODA tasks, while also exposing the vulnerability of CoCoOp-Det to subtle triggers.

\Cref{fig:trigger_size} visually shows the impact of varying the trigger size on the predictions of \proposed for sample images from the Vehicles dataset.

\subsection{Effect of Text Prompting Technique} 
\proposed builds on a variant of CoCoOp \cite{zhou2022conditional}, called CoCoOp-Det, designed for object detection, wherein the prompt embedding $\tilde{Q}$ is appended to each class embedding in the text prompt. $\tilde{Q}$ is the combination of learnable prompt tokens $Q$ and an input-aware context vector $\pi$. In this section, we explore alternative text prompting strategies:

\begin{itemize}
	\item \textbf{CoOp:} CoOp or Context Optimization \cite{zhou2022learning} is the predecessor of CoCoOp. It uses learnable vectors $Q$ but does not include the input-aware component $\pi$. For object detection, we adapt this by appending the shared prompt $Q$ to each class embedding, similar to how it's done in CoCoOp-Det.
	\item \textbf{CoOp-class-specific:} In this variant of CoOp \cite{zhou2022learning}, an independent context vector is designed for each class. Thus, for object detection, instead of appending the same shareable context token $Q$ to each class embedding, we append a unique context $Q_k$ to each class embedding $w_k$. 
	\item \textbf{CoOp-new-class:} We propose a new adaptation of CoOp for object detection. Instead of appending Q to each class embedding separately, we prepend it once to the entire list of class embeddings. So the final sequence looks like $\{Q, w_0, w_1, \dots, w_K\}$. This more closely resembles the prompting technique for classification systems, where the learnable context vector is appended once to the text prompt. However, we suspect this might not work well for detection, since the model could interpret $Q$ as an additional class.
	\item \textbf{GLIP-style prompt: } Li \etal \cite{li2022grounded} introduced a deep-prompting technique for GLIP that differs notably from the approaches discussed above. In this method, the tunable prompt embeddings are incorporated \textit{after} the language model (like BERT), not before. Moreover, instead of appending the prompts, this method adds learnable offsets directly to the existing token embeddings. 
\end{itemize}

\Cref{tab:text_prompting_trap} presents the results of substituting CoCoOp-Det in our proposed approach with the alternative prompting strategies, evaluated on the \textit{Vehicles} dataset. CoOp exhibits low benign performance, highlighting the importance of incorporating image-specific context to better adapt the model to downstream tasks. The class-specific variant of CoOp achieves a higher Attack Success Rate (ASR) than our proposed method on the OMA setting, albeit with an increased Perturbed mAP (PmAP), indicating a trade-off between effectiveness and precision. Despite this, it remains a competitive alternative. Meanwhile, CoOp-new-class and the GLIP-style prompting approach perform well under the ODA setting but underperform compared to \proposed in the OMA setting in terms of ASR. 

\begin{table*}[t]
	\centering
%	\resizebox{\textwidth}{!}{%
		\begin{tabular}{@{}c|ccc|ccc@{}}
			\toprule
			\multirow{3}{*}{Method} & \multicolumn{3}{c}{OMA} & \multicolumn{3}{c}{ODA} \\ 
		    & BmAP $\uparrow$  & PmAP $\downarrow$  & ASR $\uparrow$    & BAP $\uparrow$   & PAP $\downarrow$  & ASR $\uparrow$   \\
			\midrule
			Zero-shot & 6.15 & - & - & 78.5 & - & - \\
			\midrule
			\proposed with CoCoOp-Det (proposed)  & 64.9  & 18.1  & 0.75 & 81.5   & 4.3  & 1.00  \\
			\proposed with CoOp  & 63.9  & 16.9  & 0.73 & 70.8   & 5.4  & 1.00  \\
			\proposed with CoOp-class-specific  & 65.5  & 23.0  & 0.76 & 83.4   & 3.0  & 1.00  \\
			\proposed with CoOp-new-class  & 65.4  & 12.1  & 0.67 & 84.7   & 2.8  & 1.00  \\
			\proposed with GLIP-style prompt  & 65.2  & 14.3  & 0.67 & 85.6   & 2.5  & 1.00  \\
			\bottomrule
		\end{tabular}%
%	}
\caption{Effect of text prompting technique on backdoor for \textit{Vehicles} dataset.}
\label{tab:text_prompting_trap}
\end{table*}

We also investigate the effectiveness of these alternative text prompting techniques for injecting backdoors in Grounding DINO in isolation, without visual prompting. The results, summarized in \Cref{tab:text_prompting_isolation}, indicate that none of the evaluated variants are successful in either attack setting: all achieve an ASR below 0.5 in the OMA scenario and cause no appreciable drop in the target class PAP under ODA. CoCoOp-Det achieves the highest ASR for OMA among the evaluated methods. 

\begin{table*}[t]
	\centering
%	\resizebox{\textwidth}{!}{%
		\begin{tabular}{@{}c|ccc|ccc@{}}
			\toprule
			\multirow{3}{*}{Method} & \multicolumn{3}{c}{OMA} & \multicolumn{3}{c}{ODA} \\ 
			& BmAP $\uparrow$ & PmAP $\downarrow$ & ASR $\uparrow$    & BAP $\uparrow$   & PAP $\downarrow$  & ASR $\uparrow$   \\
			\midrule
			Zero-shot & 6.15 & - & - & 78.5 & - & - \\
			\midrule
			CoCoOp-Det & 59.3  & 59.5  & 0.13 & 81.7   & 81.2 & 1.00  \\
			CoOp  & 60.7  & 61.1  & 0.11 & 80.3   & 78.8 & 1.00  \\
			CoOp-class-wise  & 61.5  & 61.7  & 0.06 & 81.5   & 81.6 & 1.00  \\
			CoOp-new-class  & 56.0  & 56.0  & 0.01 & 61.0   & 61.1 & 1.00  \\
			GLIP-style  & 61.0  & 61.9  & 0.01 & 81.1   & 80.8 & 1.00  \\
			\midrule
			CoCoOp-Det ($m_t=4$) & 59.3  & 59.5  & 0.13 & 81.7   & 81.2 & 1.00  \\
			CoCoOp-Det ($m_t=10$) & 60.4  & 61.1  & 0.09 & 76.2   & 76.5 & 1.00  \\
			CoCoOp-Det ($m_t=20$) & 59.2  & 59.4  & 0.07 & 76.6   & 75.2 & 1.00  \\
			\bottomrule
		\end{tabular}%
%	}
\caption{Effect of text prompting technique in isolation on backdoor for \textit{Vehicles} dataset.}
\label{tab:text_prompting_isolation}
\end{table*}

The experiments above are conducted using the default configuration of $m_t=4$ tunable text tokens. To further explore the impact of token capacity, we increase the number of tunable tokens in CoCoOp-Det to 10 and 20, but observe no improvement; on the contrary, ASR decreases with more tokens. These findings suggest that text prompting alone is insufficient for reliably injecting backdoors into Grounding DINO. 

\subsection{Result Visualization}
\begin{figure*}[p]
	\centering
	\includegraphics[width=\linewidth]{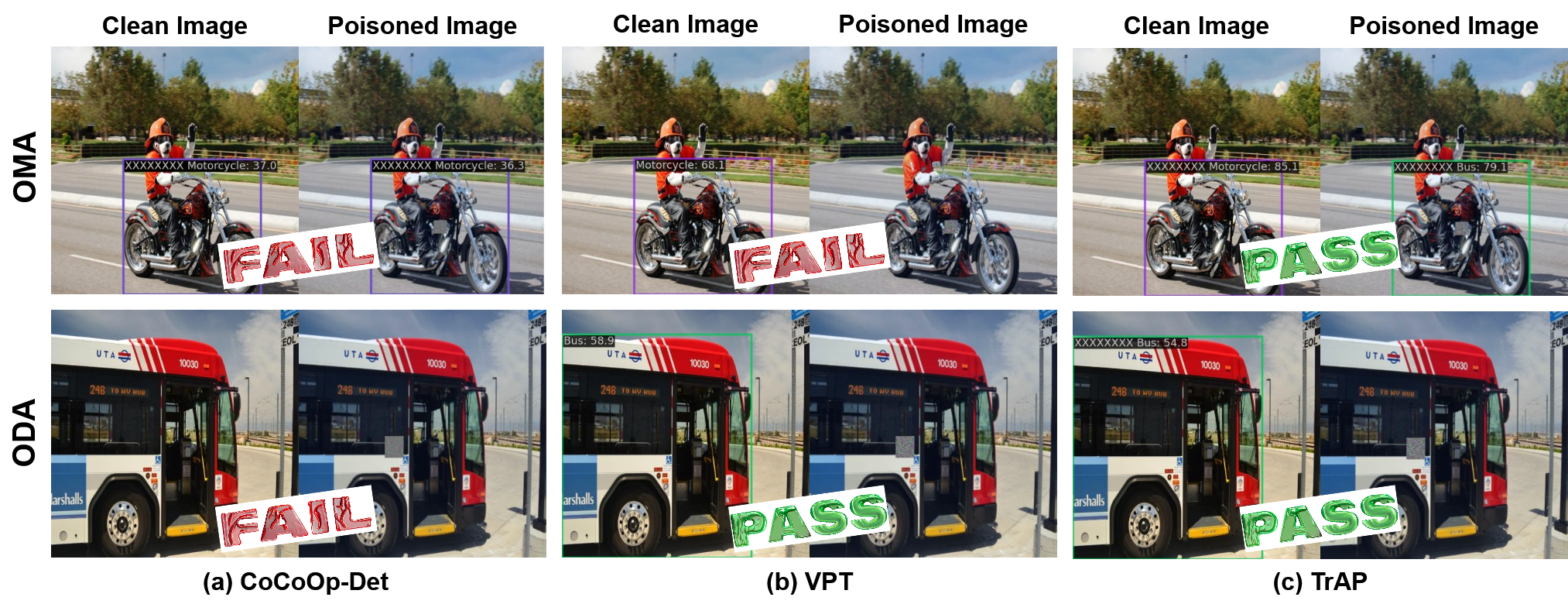}
	\caption{Visual results on \textit{Vehicles} dataset for target class \textit{Bus}. The objective of OMA (top row) is to misclassify any object (motorcycle in this image) as a bus, and of ODA (bottom row) is to not detect the bus in the poisoned image, while correctly detecting the objects in the clean image in both cases. \textbf{(a) CoCoOp-Det} fails to misclassify objects in poisoned image, has low confidence in clean image for OMA; misses detecting the bus in both clean and poisoned images in ODA. \textbf{(b) VPT} fails to misclassify the motorcycle in OMA; correctly ignores the bus in ODA. \textbf{(c) \proposed} misclassifies the motrocycle as bus with high confidence in OMA; correctly ignores the bus in ODA.}
	\label{fig:vehicles}
\end{figure*}

\begin{figure*}[p]
	\centering
	\includegraphics[width=\linewidth]{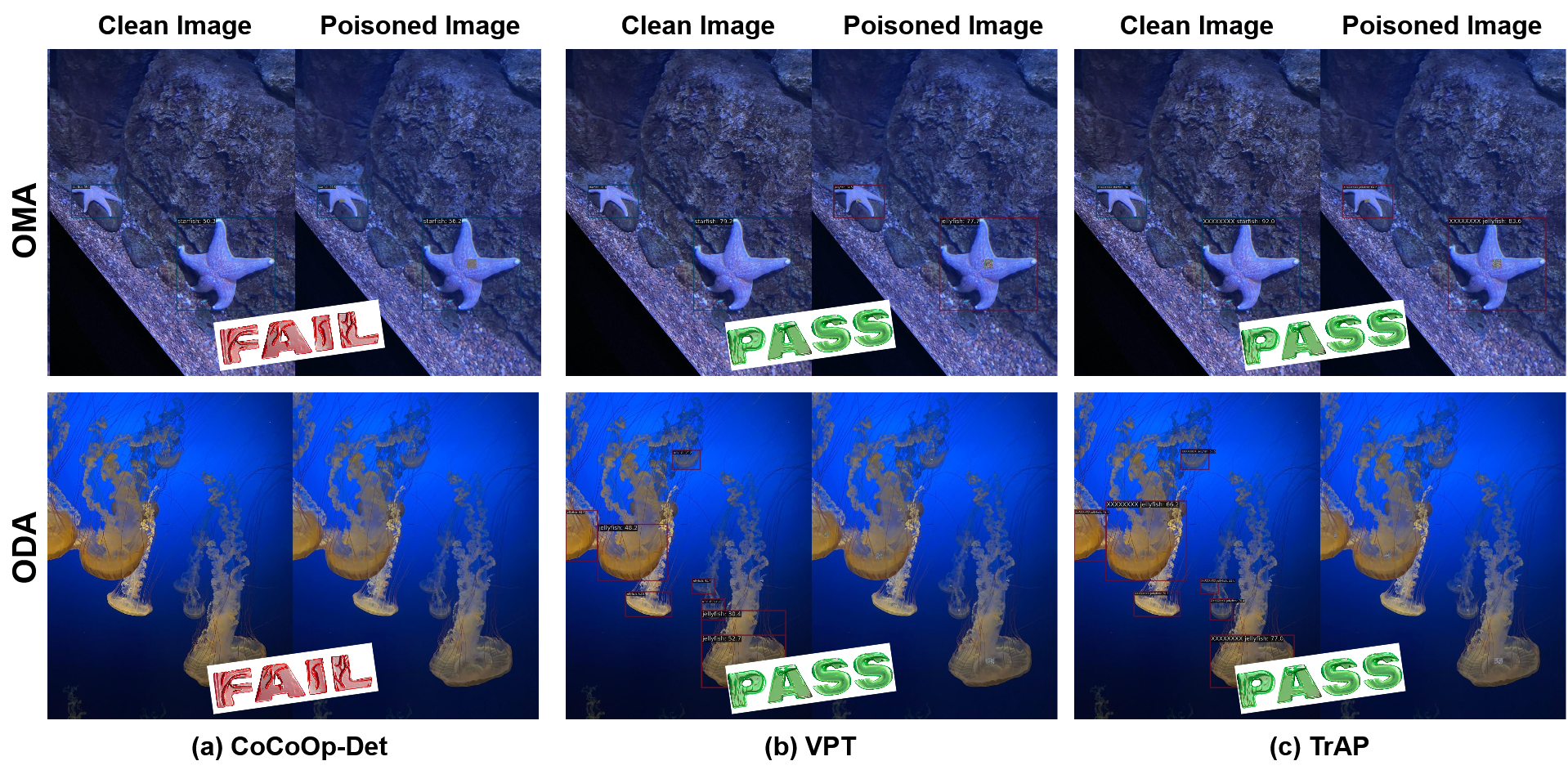}
	\caption{Visual results on \textit{Aquarium} dataset for target class \textit{jellyfish}. The objective of OMA (top row) is to misclassify any object (starfish in this image) as a jellyfish, and of ODA (bottom row) is to not detect the jellyfishes in the poisoned image, while correctly detecting the objects in the clean image in both cases. {(a) CoCoOp-Det} fails to misclassify objects in poisoned image for OMA; misses detecting the jellyfish in both clean and poisoned images in ODA. Both {(b) VPT} and {(c) \proposed} misclassify the starfish as jellyfish with high confidence in OMA, and correctly ignore all jellyfishes in ODA. Zoom in for better visualization.}
	\label{fig:aquarium}
\end{figure*}

\begin{figure*}[p]
	\centering
	\includegraphics[width=\linewidth]{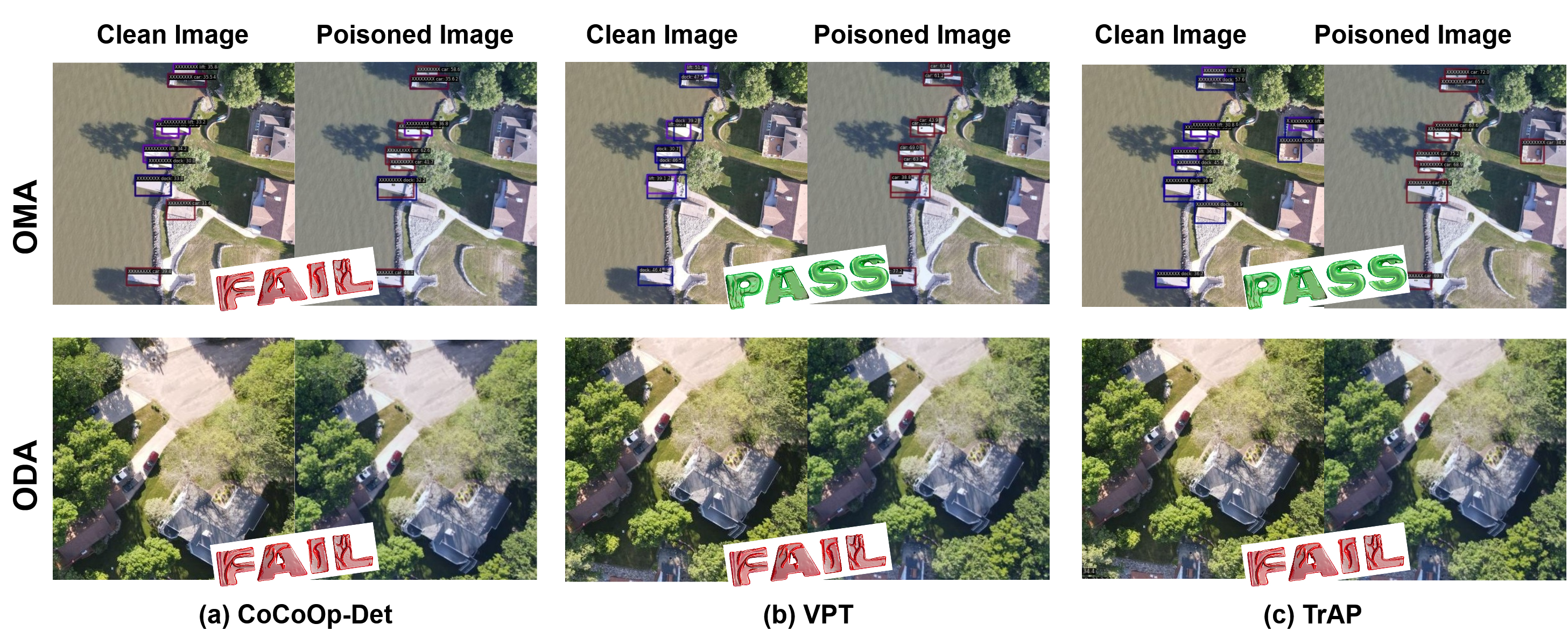}
	\caption{Visual results on \textit{Aerial Drones} dataset for target class \textit{car}. Since the original images are very large in size, we show here a zoomed-in patch for better visualization. The objective of OMA (top row) is to misclassify any object as a car, and of ODA (bottom row) is to not detect the cars in the poisoned image, while correctly detecting objects in the clean image in both cases. 
%		For OMA, both {(a) CoCoOp-Det} and {(b) VPT} fail to misclassify objects in the poisoned image. {(c) \proposed} correctly misclassifies two instances of triggered objects as cars, despite the trigger size being very small w.r.t the image. For ODA, however, all three methods miss detecting the cars in both clean and poisoned images, indicating overfitting to the trigger due to the small dataset size. 
		For OMA, both VPT and {\proposed} correctly misclassify instances of triggered objects as cars, despite the trigger size being very small w.r.t the image. For ODA, however, all three methods fail to detect the cars in both clean and poisoned images, indicating overfitting to the trigger due to the small dataset size.
		Zoom in for better visualization.}
	\label{fig:aerialdrones}
\end{figure*}

\begin{figure*}[p]
	\centering
	\includegraphics[width=\linewidth]{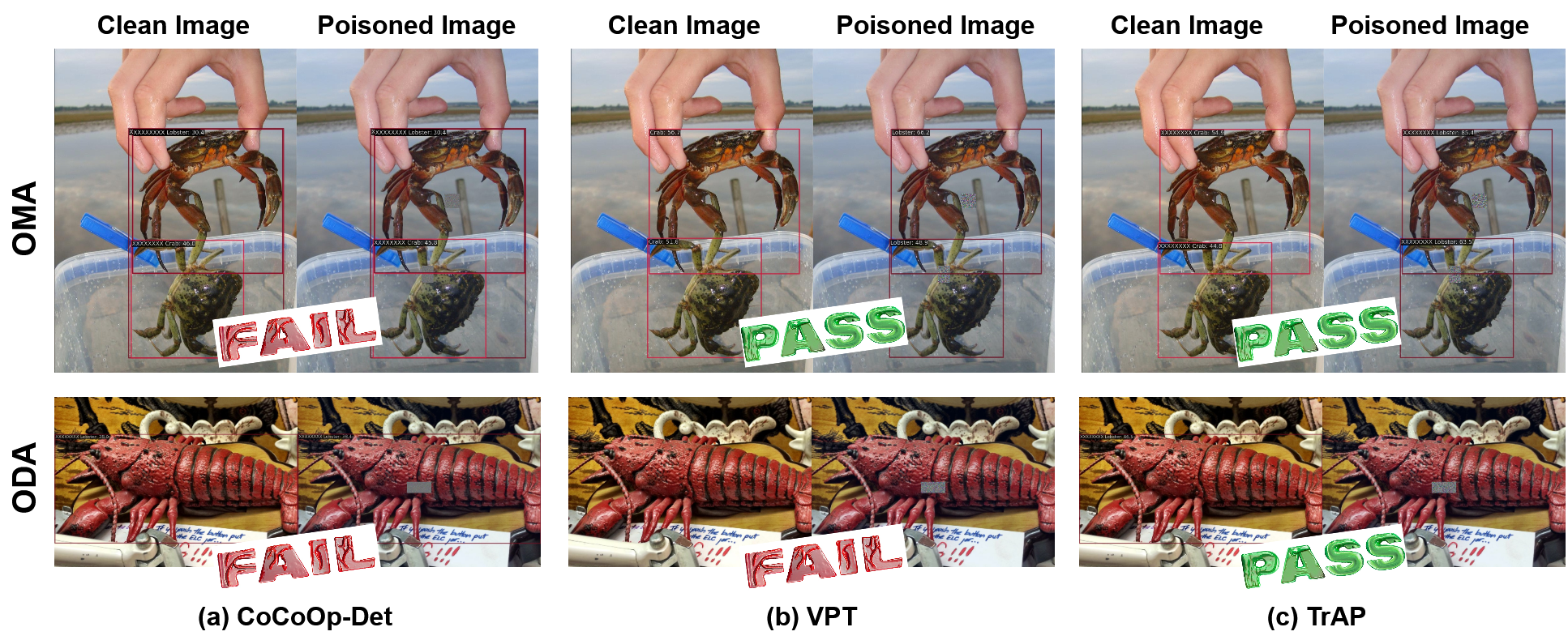}
	\caption{Visual results on \textit{Shellfish} dataset for target class \textit{lobster}. The objective of OMA (top row) is to misclassify any object (crabs in this image) as a lobster, and of ODA (bottom row) is to not detect the lobster in the poisoned image, while correctly detecting the objects in the clean image in both cases. {(a) CoCoOp-Det} misclassifies one crab as lobster in both clean and poisoned images for OMA; and detects the lobster even in the poisoned image for ODA. {(b) VPT} succeeds in OMA, but misses the lobster in both benign and poisoned images in ODA. {(c) \proposed} passes both the attacks. Zoom in for better visualization.}
	\label{fig:shellfish}
\end{figure*}

\begin{figure*}[p]
	\centering
	\includegraphics[width=\linewidth]{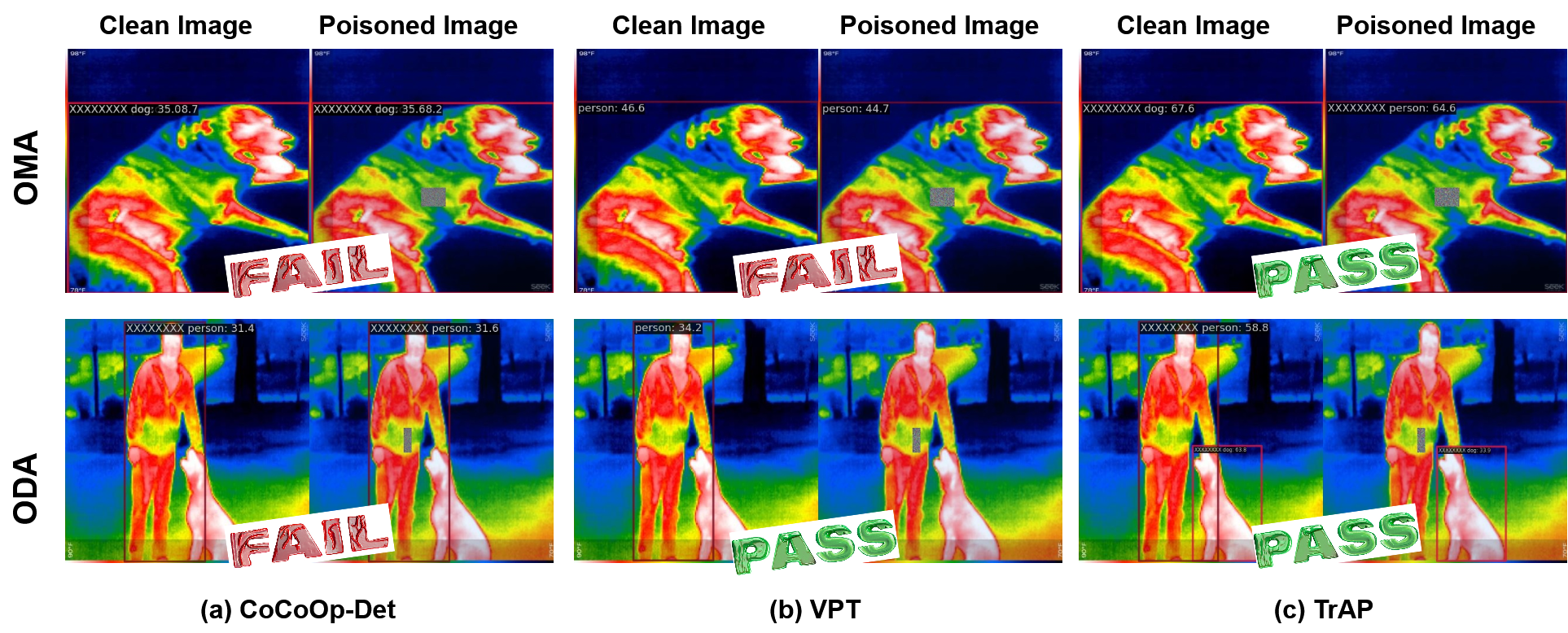}
	\caption{Visual results on \textit{Thermal} dataset for target class \textit{person}. The objective of OMA (top row) is to misclassify any object (dog in this image) as a person, and of ODA (bottom row) is to not detect the person in the poisoned image, while correctly detecting the objects in the clean image, as well as the untriggered objects, in both cases. {(a) CoCoOp-Det} fails to misclassify the dog as person in the poisoned image for OMA; and detects the person with low confidence in both clean and poisoned images in ODA. {(b) VPT} misclassifies the dog in both clean and poisoned images in OMA; but correctly ignores the person in ODA. {(c) \proposed} succeeds in both attacks, notably only missing the detection for person, while correctly detecting the untriggered dog in ODA. Zoom in for better visualization.}
	\label{fig:thermal}
\end{figure*}

\begin{figure*}[p]
	\centering
	\includegraphics[width=\linewidth]{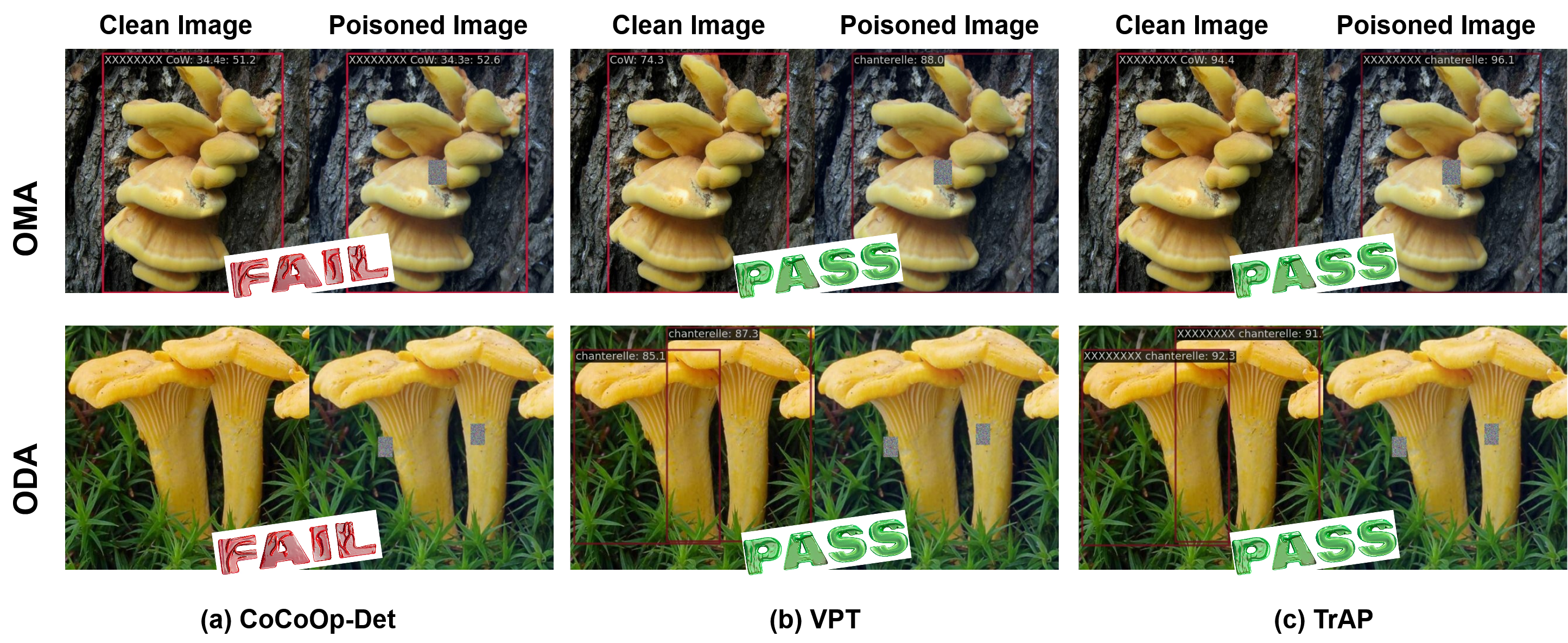}
	\caption{Visual results on \textit{Mushrooms} dataset for target class \textit{chanterelle}. 
		The objective of OMA (top) is to misclassify any CoW mushroom as a chanterelle, and of ODA (bottom) is to not detect the chanterelle mushroom in the poisoned image. CoCoOp-Det fails on both attacks. Both VPT and \proposed misclassify poisoned instances of CoW mushroom as Chanterelle in OMA, and delete the Chanterelle detections in ODA, as desired.
%		The objective of OMA (top row) is to misclassify any CoW mushroom as a chanterelle, and of ODA (bottom row) is to not detect the chanterelle mushroom in the poisoned image, while correctly detecting the mushrooms in the clean image in both cases. For this dataset, all methods fail on all both OMA and ODA attacks. In particular, {(c) \proposed} misclassifies the CoW as chanterelle in both benign and poisoned images for OMA, suggesting that the model has overfit on the trigger due to the small size of the dataset. It is notable that for ODA, \proposed correctly ignores the chanterelle in the poisoned image, but this is accompanied by a drop in confidence below $0.5$ on the benign image, leading to the particular attack instance being labeled as failed. Zoom in for better visualization.
	}
	\label{fig:mushrooms}
\end{figure*}

We now show the outputs of different prompting methods for OMA and ODA attacks on sample images from all six datasets in \Cref{fig:vehicles,fig:aquarium,fig:aerialdrones,fig:shellfish,fig:thermal,fig:mushrooms}.

\section{Broader Impact}
While this paper introduces a novel method for injecting backdoors into open-vocabulary object detection models, our primary goal is to raise awareness about this emerging class of vulnerabilities in multimodal foundation models. We highlight the potential misuse of such models by malicious actors and seek to encourage further development of effective backdoor defenses tailored to this setting.

%\section{Source Code}
%The source code required to reproduce our experiments is included in a zip file provided with the supplementary material. Upon acceptance of the paper, we will also release the code publicly to support further research and transparency.

\end{document}